\documentclass{article} 
\usepackage{iclr2027_conference,times}

\usepackage{amsmath,amsfonts,bm}

\def\eqref#1{equation~\ref{#1}}

\def\1{\bm{1}}

\DeclareMathAlphabet{\mathsfit}{\encodingdefault}{\sfdefault}{m}{sl}
\SetMathAlphabet{\mathsfit}{bold}{\encodingdefault}{\sfdefault}{bx}{n}

\usepackage{hyperref}
\usepackage{url}
\usepackage{booktabs}
\usepackage{graphicx}
\usepackage{xcolor}
\usepackage{booktabs}

\usepackage{colortbl}

\title{ReMoVe: Refining a Video Model's Own Motion for Coherent Human Video Generation}

\author{
Jie Li$^{1,2}$\thanks{Work done during internship at SJTU.},
Muxin Pu$^{3}$,
Hongyi Cai$^{4}$,
Mingkang Dong$^{5}$,
Jianyuan Guo$^{6}$,
Tao Huang$^{1}$\thanks{Corresponding author.}
\\[6pt]
$^{1}$Shanghai Jiao Tong University \quad
$^{2}$University of Science and Technology Beijing \quad
$^{3}$Monash University \\
$^{4}$Universiti Malaya \quad
$^{5}$The University of Queensland \quad
$^{6}$City University of Hong Kong \\
}

\newcommand{\hlbest}[1]{\cellcolor{orange!45}#1}
\newcommand{\hlsecond}[1]{\cellcolor{orange!18}#1}
\iclrfinalcopy
\begin{document}

\maketitle
\begin{figure}[htbp]
\centering
\includegraphics[width=\linewidth]{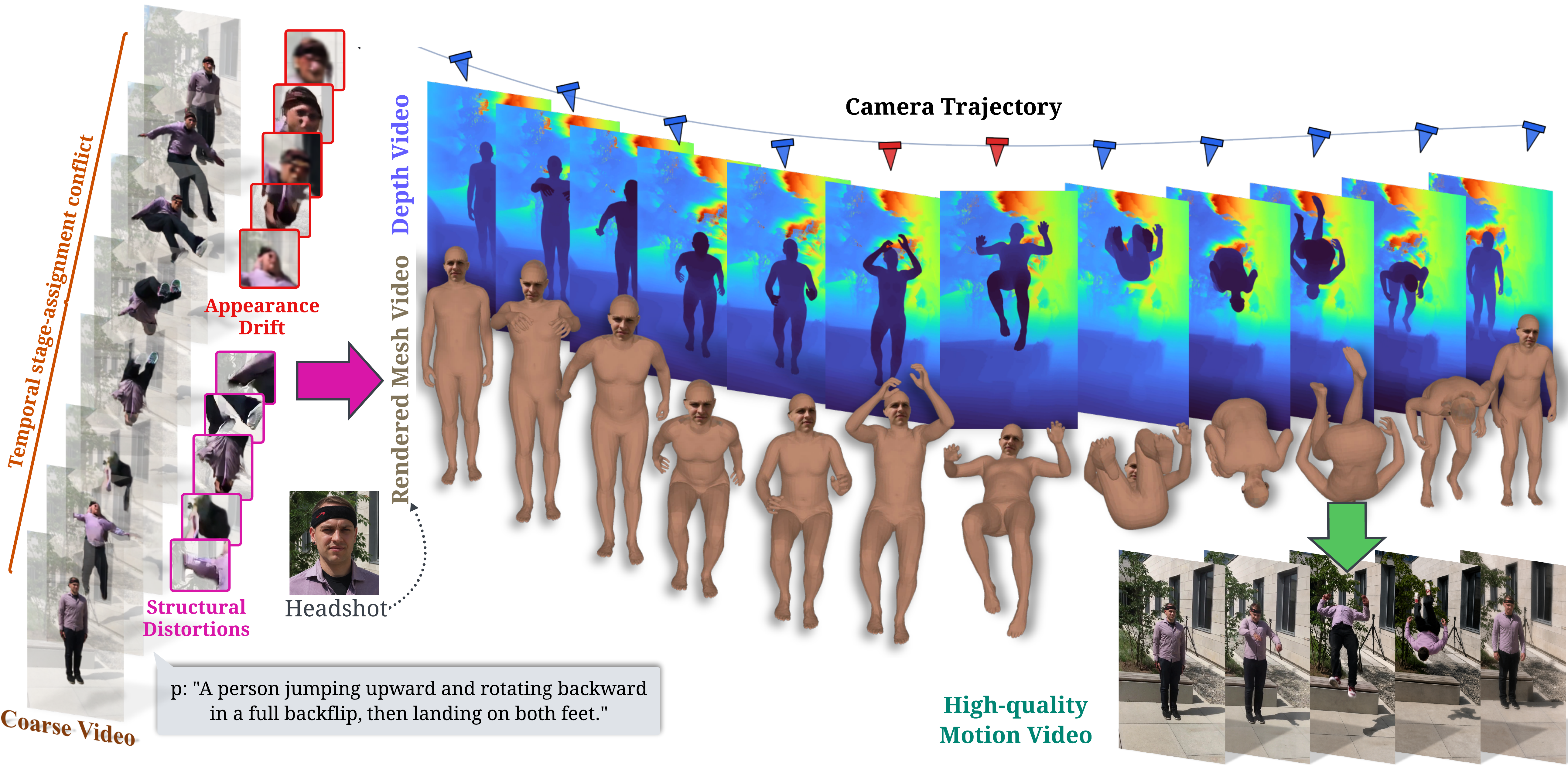}
\caption{Starting from a coarse video produced by a base video generator, together with a headshot from its initial frame, we recover and refine the underlying motion, render a paired mesh video with corresponding depth, and regenerate a temporally coherent, identity-consistent video under a user-specified camera trajectory.}
\label{fig:teaser}
\end{figure}

\begin{abstract}

Video generation models have become increasingly capable of synthesizing realistic human motion, yet they often lose track of the very structure that makes the motion plausible. As generated human motion becomes more complex, videos can suffer from progressive body deformation and appearance drift across frames. Existing motion-controlled methods use explicit representations to stabilize generation, but typically rely on pre-defined motion sequences as references. This raises a fundamental question: if the base generator already produces a usable coarse motion, why replace it with an external one? \textbf{ReMoVe} answers this question by treating the coarse output as a motion hypothesis rather than a final result. ReMoVe recovers and refines the underlying motion, then renders the refined motion as 3D mesh and transfers facial appearance onto it, thereby improving motion accuracy and identity consistency. We further derive depth from the same 3D scene and camera trajectory to impose scene-level spatial consistency beyond the mesh.
By jointly conditioning video generation models on mesh and depth, ReMoVe synthesizes high-quality motion videos with controllable camera trajectories and enhanced spatio-temporal consistency. 
Extensive experiments demonstrate that ReMoVe improves human structural stability, appearance consistency, temporal coherence, and motion-following fidelity, while retaining the base generator's motion prior and offering accurate camera control.
All open-source project assets are now publicly available at \url{https://remove-video.github.io}.

\end{abstract}

\section{Introduction}

Recent advances in video generation models have substantially improved visual fidelity and text alignment, enabling increasingly expressive generated videos~\citep{wan2025,yang2024cogvideox,kong2024hunyuanvideo}.
Building on these advances, recent works have increasingly explored human video generation~\citep{zhao2026comovi,cao2025uni3c}. However, existing methods still struggle to maintain motion accuracy, physical consistency, and appearance consistency, particularly during large-amplitude movement~\citep{fang2026motioncharacterfinegrainedmotioncontrollable,tu2025stableanimator}. 
As shown in the left side of Fig.~\ref{fig:teaser}, two failure modes can be observed: structural distortion (e.g., progressive deformation and collapse of limbs into indistinguishable blobs) and appearance drift (e.g., inconsistent face identity across frames).

These failure modes are often intertwined: temporal inconsistencies in motion can lead to anatomically implausible poses and further compromise the subject's appearance across multiple frames. 
We further observe that this problem becomes more pronounced for complex, goal-directed actions such as backflips and cartwheels, which involve rapid pose transitions and require precise coordination across long temporal spans~\citep{taghipour2026controllablecomplexhumanmotion}. 
Since generators synthesize a fixed number of frames without any explicit stage-wise supervision, they may compress or skip intermediate transitions when the temporal budget is insufficient, yielding unnatural motion, as illustrated by the backflip example in Fig.~\ref{fig:teaser}, where the rotation collapses into a single implausible jump. 
This is not merely a data-scaling problem. Commercial models have demonstrated the ability to synthesize increasingly realistic and complex human motion videos within a limited frame budget.
Even when the result appears relatively natural, the generated motion can exceed the physical limits of real-world human motion. 
We refer to this mismatch as a \textbf{temporal stage-assignment conflict}, and validate it empirically in Appendix~\ref{sec:appendix_temporal_conflict}.

This issue is further compounded by the lack of explicit constraints on human body structure in current generators: without an anatomical prior to fall back on, an abrupt transition can result in anatomically implausible poses rather than simply faster movement. 
Since motion and appearance are jointly synthesized over the evolving body structure, such pose distortions can further propagate to body-attached appearance, including clothing, hair, and facial identity.
Moreover, conditioning on a single reference-image embedding conveys the subject's identity from only one viewpoint, but provides limited information about how that identity should be preserved under new viewpoints. As large-amplitude articulation continuously changes the viewpoint, facial identity tends to drift.
Existing methods address these issues by providing an additional motion sequence, such as 2D skeletons or text-to-motion outputs~\citep{hu2025animate,wang2025humandreamer}; using parametric body models or dense renderings to constrain geometry and pose~\citep{zhu2024champ}; or conditioning on a limited set of reference images to preserve identity~\citep{zhang2025fantasyid}.

However, these pipelines suffer from two complementary limitations. 
First, methods that rely on an external motion sequence assume the desired motion is given in advance, either extracted from an existing video~\citep{ding2025mtvcrafter}, which requires the target action to be captured beforehand, or synthesized by a text-to-motion model~\citep{wang2025humandreamer}. 
Such reliance limits their scalability and may produce motion that is disconnected from the generated scene. 
Second, methods that condition on a limited set of reference images provide a stable identity signal but no information about how the subject should appear under the diverse viewpoints induced by motion and camera trajectory, so identity still drifts as articulation changes the viewpoint.

In this paper, we propose \textbf{ReMoVe}, a self-bootstrapping conditioning pipeline that treats the base video generator's coarse output as a motion hypothesis already encoding a plausible action and scene interaction. 
Reliable segments of the recovered motion are retained as anchors, and a keyframe-conditioned motion generation model completes the unreliable spans around them. This allows the refined motion to inherit the base generator's original motion prior and scene grounding, rather than being regenerated from scratch or constrained to a fixed frame length. 
To turn this motion into a conditioning signal that a video generator can directly follow, ReMoVe renders it as a mesh video, placing the target identity's facial appearance on a posed SMPL-X mesh at the appropriate viewpoint in each frame. This converts structural and appearance correction into an explicit, pixel-space signal rather than a numerical pose parameter or a single 
reference-image embedding.

Besides imposing physical constraints on the motion, the mesh video alone leaves the camera trajectory and background changes to the generator's own inference, which becomes unstable as the action grows more complex. We therefore render the mesh and depth videos from the same 3D scene using a shared camera trajectory, ensuring that the two conditions remain geometrically consistent. 
This allows subject motion and camera motion to be controlled independently while remaining geometrically consistent, unlike standard text-conditioned generators, where both are implicitly encoded in a single prompt, or camera-trajectory controllers that lack an explicit condition for articulated human motion~\citep{he2025cameractrliidynamicscene}.
Across both signals, the underlying principle is the same. 
Rather than asking the video generator to implicitly reason about precise body kinematics and camera geometry, we resolve them explicitly through motion refinement and 3D rendering, allowing the generator to focus on what it already does well, synthesizing appearance conditioned on a fully specified structure.

Extensive experiments validate the efficacy of ReMoVe. We conduct comprehensive evaluations on the Motion-X++ dataset~\citep{zhang2025motion} using the VBench~\citep{huang2023vbench} and VBench-2.0~\citep{zheng2025vbench2} benchmarks, together with distributional and reconstruction metrics.
Qualitative and quantitative results show that ReMoVe improves structural stability, appearance consistency, and motion controllability over base video generators and existing motion-aware and motion-conditioned baselines, while preserving the base video generator's motion prior. 
We further perform ablation studies on the mesh and depth conditions, demonstrating their complementary contributions to the final performance. Finally, evaluations on camera-controlled video generation show that ReMoVe enables accurate and independent control of both human motion and camera motion. 

\section{Related Works}
\label{related}

\subsection{Motion-Controlled Video Generation}

Explicit motion conditioning has been widely explored to improve controllability and structural stability, typically via 2D keypoints, parametric human models, depth maps, or rendered structural representations. One line of work builds a complete motion sequence directly within or alongside the generative process: 
VideoJAM~\citep{chefer2025videojamjointappearancemotionrepresentations} augments a pretrained video model to jointly predict RGB appearance and optical flow, EchoMotion~\citep{yang2025echomotion} jointly denoises video tokens and parametric human motion in a dual-modality diffusion transformer, and CoMoVi~\citep{zhao2026comovi} couples motion and video generation in a single denoising loop through a bridging 2D representation.
Other efforts strengthen appearance or geometric constraints rather than motion itself: identity conditioning has moved from single-image portraits to 3D facial priors and multi-view full-body references ~\citep{xu2025hunyuanportraitimplicitconditioncontrol,guo2026wildactor}, 
and some works align reconstructed scene geometry with a posed human body for joint camera and motion control~\citep{cao2025uni3c}.

Across these directions, motion is usually generated separately from the scene, subject, and camera in the video, or the video model itself has to be redesigned to learn motion and video together. Neither approach actually fixes the motion errors that come out of a single generation.
Closest to our setting, ReVision~\citep{liu2026revisionrefiningvideodiffusion} shows that a coarse video generation can itself provide a useful motion hypothesis, iteratively refining a 3D motion representation recovered from it before feeding the result back for regeneration, yet this feedback mainly strengthens motion consistency, leaving identity appearance and scene-level geometry largely delegated to the video generator.
Motion coherence, identity consistency, and scene evolution thus appear closely coupled under large and complex human motions. We instead aim to preserve the motion and scene priors already present in a pretrained video generator, correct only the unreliable motion states, and supply geometrically aligned appearance and scene constraints during regeneration.

\subsection{Human Motion Generation}

Human motion generation has progressed from early RNN-, VAE-, and GAN-based approaches to Transformer and diffusion-based models supporting flexible, controllable synthesis~\citep{tevet2022humanmotiondiffusionmodel,cohan2024flexiblemotioninbetweeningdiffusion,zhang2023t2mgptgeneratinghumanmotion}.
More recently, large-scale motion foundation models have emerged. HY-Motion~\citep{wen2025hymotion10scalingflow} scales DiT-based flow matching to billions of parameters, while Kimodo~\citep{rempe2026kimodoscalingcontrollablehuman} combines motion diffusion with kinematic control beyond text-only conditions.
However, these models primarily synthesize motion from semantic descriptions, without accounting for the initial pose, identity, or scene context of a specific video. We thus use them as motion priors for refinement rather than direct control signals, as detailed in Sec.~\ref{method}.

\section{Method}
\label{method}

\begin{figure}[t]
\centering
\includegraphics[width=\linewidth]{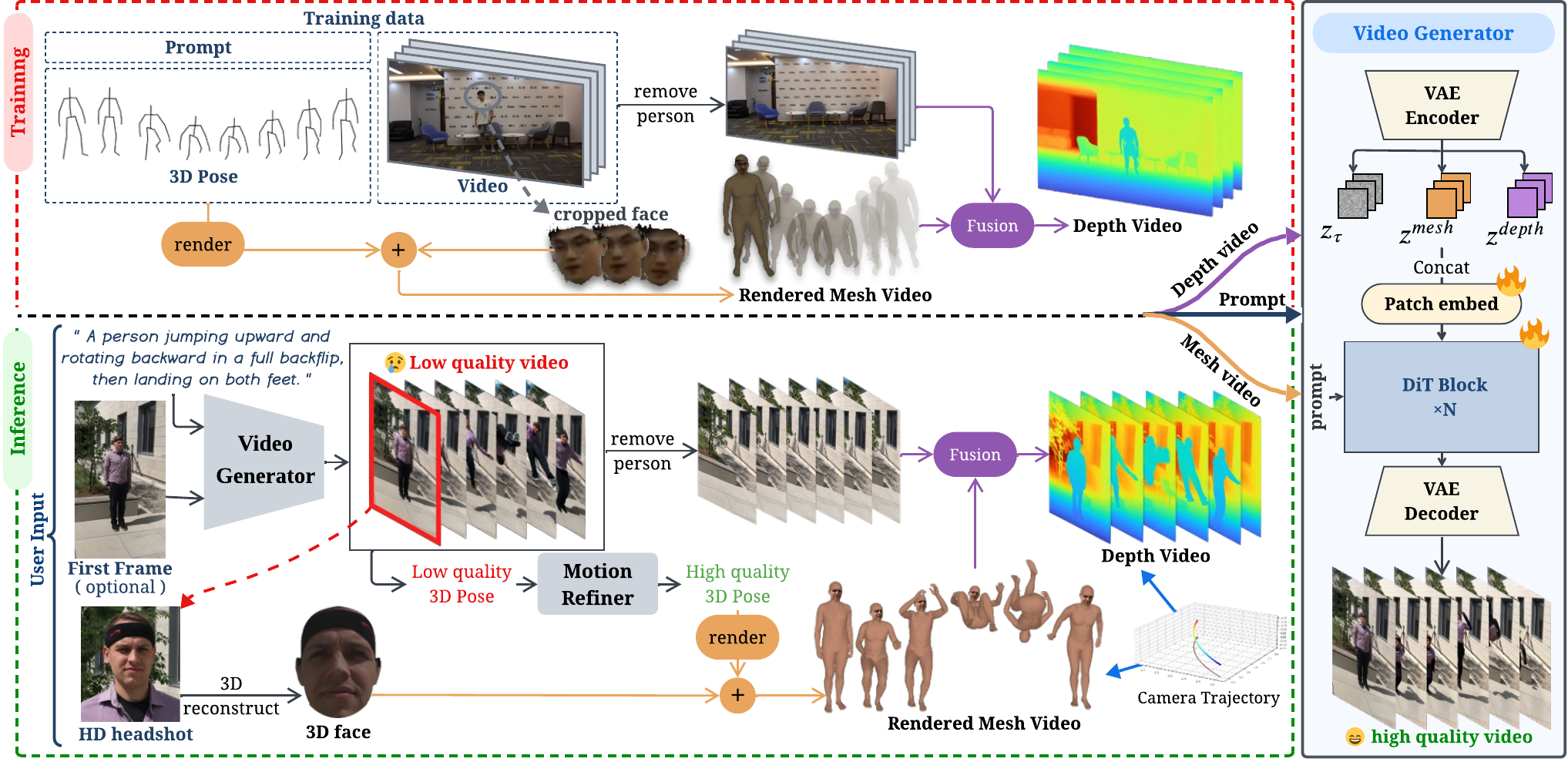}
\caption{Overview of ReMoVe. On the left is the construction of mesh and depth conditions during training (top) and inference (bottom). On the right is the generation pipeline, where the mesh and depth videos are encoded by the VAE, concatenated with the noisy video latent, and fed into the DiT together with the text prompt to produce the final high-quality video.}
\label{fig:method}
\end{figure}

\subsection{Overview}
\label{sec:overview}

Our goal is to improve human structural stability, identity consistency, and motion control while preserving the base video generator's motion prior.
It conditions the generator on aligned mesh and depth videos rendered under a shared camera trajectory and injected at the latent level.
During training, these conditions are constructed from ground-truth annotations; at inference, they are instead reconstructed from the base model's own coarse generation after motion refinement. The resulting representations follow the same form in both cases, as described in Sec.~\ref{sec:training_data} and Sec.~\ref{sec:inference_cond} respectively.

\subsection{Conditioning Representations}
\label{sec:conditioning}

For a target video of $T$ frames, let $\theta_t$ denote the human pose parameters at frame $t$ and $\mathcal{M}_t$ the corresponding posed parametric mesh in the camera frame.
We drive only the body pose because Kimodo~\citep{rempe2026kimodoscalingcontrollablehuman}, our inference-time motion source, outputs only SMPL body pose. Hand articulation, jaw, and facial expression are therefore fixed to a canonical rest state during both training and inference to keep the rendered conditions consistent.
From $\{\mathcal{M}_t\}_{t=1}^{T}$ and a target camera trajectory, we render the frame-aligned mesh condition $C^{\mathrm{mesh}}$ and depth condition $C^{\mathrm{depth}}$.

\paragraph{Rendered mesh conditioning.}

The rendered mesh conditioning video
$C^{\mathrm{mesh}}=\{C_t^{\mathrm{mesh}}\}_{t=1}^{T}$
is obtained by rasterizing the posed mesh $\mathcal{M}_t$ into the target camera view and rendering its surface with two complementary appearance regions. The facial region is assigned the appearance of the target identity, providing explicit identity and facial appearance signals. The remaining body region is rendered with a single flat skin tone modulated by surface-normal-based shading, so that body shape and pose are conveyed primarily through silhouette and geometric shading without introducing specific clothing or texture patterns. 

\paragraph{Depth conditioning.}

The depth conditioning video $C^{\mathrm{depth}}=\{C_t^{\mathrm{depth}}\}_{t=1}^{T}$ encodes scene geometry and the spatial placement of the target person under the corresponding camera view, including relative depth, occlusion relationships, and global spatial layout. It is rendered under the same camera trajectory as $C^{\mathrm{mesh}}$, establishing a frame-wise correspondence between human motion, 
scene geometry, and viewpoint changes.

\subsection{Training Data Construction}
\label{sec:training_data}

Each training clip consists of a real captured video $V=\{I_t\}_{t=1}^{T}$ together with per-frame ground-truth SMPL-X motion parameters $\Theta=\{\theta_t\}_{t=1}^{T}$ and camera parameters, temporally and spatially aligned with $V$.
From these, we construct a rendered mesh conditioning video $C^{mesh}=\{C^{mesh}_t\}_{t=1}^{T}$ and a depth conditioning video $C^{depth}=\{C^{depth}_t\}_{t=1}^{T}$, temporally aligned with $V$ and paired with the corresponding motion caption for model training, as detailed below.

\paragraph{Rendered mesh conditioning.}

The same posed SMPL-X mesh $\mathcal{M}_t$ is rasterized to obtain a dense per-pixel face-region mask $F_t$ using the predefined head-region faces of the mesh. For pixels within $F_t$, we directly copy the corresponding pixels from the real frame $I_t$, so the facial appearance retains the subject's exact pose, illumination, and identity at that frame.
For the remaining body region, we render the posed mesh using a single flat skin color sampled from the median of the visible face pixels within the clip, together with a surface-normal-based Lambertian shading term.
Specifically, the rendered mesh conditioning frame is defined as

\begin{equation}
C^{mesh}_t
=
F_t\odot I_t
+
(1-F_t)\odot R(\mathcal{M}_t,c,N_t),
\end{equation}

where $R(\cdot)$ denotes the mesh rendering function, $c$ is the sampled skin color, and $N_t$ represents the surface normals used for Lambertian shading.

\paragraph{Depth conditioning.}

We remove the person from $V$ using SAM3~\citep{carion2025sam3segmentconcepts} to obtain a foreground mask $S_t$ for each frame.
We then stack the frames with person pixels marked invalid and take a per-pixel temporal median, recovering background pixels observed elsewhere in the clip.
Only pixels never observed are spatially inpainted.
Because this procedure assumes a largely static camera, we exclude clips with substantial camera motion during data curation.
A metric depth estimator~\citep{depthanything3} then produces $D_t^{bg}$ from the completed background.

In parallel, the ground-truth motion parameters $\Theta$ are used to pose an SMPL-X mesh $\mathcal{M}_t$, which is rasterized in the corresponding camera view to obtain both its silhouette $\Omega_t$ and its geometric depth $D_t^{person}$. 
Since the body is reconstructed in its own units while $D_t^{bg}$ is metric, the two are aligned by a single per-clip scale factor estimated from frames in which the subject is in stable ground contact.
Taking the person depth directly from the mesh, rather than compositing the mesh into the frame and re-estimating depth, keeps this identical in form to the inference-time procedure of Sec.~\ref{sec:inference_cond}. 
We define the valid mesh-person region as
\begin{equation}
A_t=\Omega_t\cap S_t,
\end{equation}
where the intersection removes mesh projections onto non-human foreground objects that occlude the person in the original video (e.g., held props). The final depth condition is obtained by fusing the person and background depth:
\begin{equation}
\label{eq:depth_fusion}
C^{depth}_t
=
A_t\odot D^{person}_t
+
(1-A_t)\odot D^{bg}_t .
\end{equation}

\subsection{Inference-Time Conditioning Construction}
\label{sec:inference_cond}
At inference time, we construct the rendered mesh and depth conditioning signals from the generated video through a dedicated pipeline consisting of five stages: (i) coarse video synthesis, (ii) motion refiner, (iii) camera trajectory, (iv) rendered mesh conditioning, and (v) depth conditioning. 

\paragraph{Coarse video synthesis.}
We first use the base video generation model to synthesize a coarse video $V_0$ from the input text prompt, optionally conditioned on a provided first frame. 
$V_0$ provides an initial, potentially imperfect motion hypothesis sampled from the motion prior of the base model. 

\paragraph{Motion refiner.}

\begin{figure}[htbp]
\centering
\includegraphics[width=\linewidth]{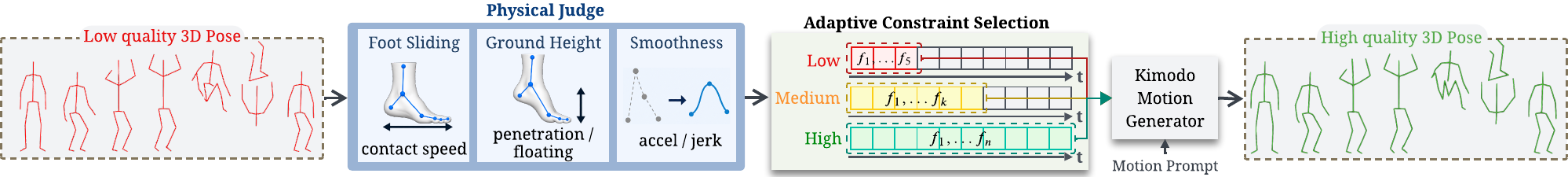}
\caption{Overview of the Motion Refiner module in ReMove.}
\label{fig:motion_refiner}
\end{figure}

As illustrated in Fig.~\ref{fig:motion_refiner}, we first project the coarse video $V_0$ into a structured human motion space using an off-the-shelf monocular human mesh recovery method, GVHMR~\citep{shen2024gvhmr}, which estimates an initial SMPL-X motion sequence $\hat{\Theta}=\{\hat{\theta}_t\}_{t=1}^{T}$. 
When $V_0$ contain visually implausible human deformation caused by generation artifacts (e.g., disconnected limbs or abrupt shape changes), fitting a parametric body model to it yields a motion sequence constrained to plausible human kinematics, filtering out much of this pixel-level noise.

Since the recovered sequence can still contain inaccurate poses or tracking failures, we further evaluate the reliability of each frame using physical-plausibility signals, retaining only reliable frames as motion anchors. Full details of these metrics and the resulting anchor-selection criteria are provided in Appendix~\ref{sec:appendix_motion_refiner}.

Let $\mathcal{K}$ denote the set of retained frame indices. We refine the motion by conditioning a text- and keyframe-guided motion synthesis model~\citep{Kimodo2026} on the retained motion anchors and the input text prompt $\mathbf{y}$, yielding a complete and temporally coherent motion trajectory
\begin{equation}
\Theta^{ref}
=
\mathcal{R}
\left(
\{\hat{\theta}_t\}_{t\in\mathcal{K}},
\mathbf{y}
\right)
=
\{\theta^{ref}_t\}_{t=1}^{T},
\end{equation}
where $\mathcal{R}(\cdot)$ denotes the motion refinement model, preserving the reliable motion content recovered from $V_0$.

\paragraph{Camera trajectory.}
The recovered motion is expressed in the per-frame camera of the coarse video, so a viewpoint change is specified as an increment on top of it. We define a relative trajectory $\mathcal{T}_{cam}=\{(R_t,\mathbf{t}_t)\}_{t=1}^{T}$, under which a point $\mathbf{x}_t$ maps to $\mathbf{x}'_t=R_t^{\top}(\mathbf{x}_t-\mathbf{t}_t)$, with the identity transform reproducing the original viewpoint. The same trajectory is applied to the posed mesh and the scene geometry, so the two conditions always correspond to a single viewpoint.

The mesh is a closed surface and can be re-rasterized exactly from any viewpoint. The scene, however, is available only as a single-layer depth map: once the camera translates, previously occluded regions become visible with no depth available and must be filled by inpainting, which limits the usable range of camera motion. Joint reprojection also requires a shared scale between the two, aligned via the same per-clip scale factor described in Sec.~\ref{sec:training_data}.

\paragraph{Rendered mesh conditioning.}
Given the refined SMPL-X motion $\Theta^{ref}$ and the target camera trajectory $\mathcal{T}_{cam}$, we pose an SMPL-X mesh $\mathcal{M}_t$ for each frame and render it under the target viewpoint, following the same face-region and shaded-body representation as in training. 
Since the coarse video $V_0$ cannot provide temporally consistent facial observations, we instead reconstruct a static FLAME texture atlas $\mathcal{A}_{\mathrm{id}}$ from a reference portrait $I_{\mathrm{ref}}$ of the target identity using a monocular face reconstruction model~\citep{retsinas2025smirk3dfacialexpressions,flame}.
For each frame, facial pixels are associated with the aligned FLAME surface and textured by barycentric sampling from $\mathcal{A}_{\mathrm{id}}$, while the remaining body region is rendered with a flat skin tone and surface-normal-based Lambertian shading, with a soft boundary between the two to avoid visible discontinuities, yielding the conditioning sequence $C^{mesh}=\{C_t^{mesh}\}_{t=1}^{T}$.

\paragraph{Depth reconstruction.}
We follow the same procedure as Sec.~\ref{sec:training_data} to remove the person from $V_0$, reconstruct the background by temporal median, and estimate its metric depth $D_t^{bg}$, with two differences introduced by the inference setting. 
First, a target camera trajectory can optionally be specified. When provided, $D_t^{bg}$ is unprojected into a point cloud, transformed according to $(R_t,\mathbf{t}_t)$, and rendered from the target viewpoint. The resulting disocclusion holes are filled through depth-domain inpainting, while $D_t^{person}$ is obtained by rasterizing the posed mesh from the same viewpoint and aligned with $D_t^{bg}$ using a shared per-clip scale factor.
Second, since the foreground mask extracted from $V_0$ is only valid in the original view, the person region under a new viewpoint is given by the rendered mesh silhouette alone, $A_t=\Omega_t$, reducing to $\Omega_t\cap S_t$ when the original viewpoint is retained. 
The final depth condition is obtained as in Eq.~\ref{eq:depth_fusion}, yielding the depth conditioning sequence $C^{depth}=\{C_t^{depth}\}_{t=1}^{T}$.

\subsection{Model Architecture}
\label{sec:architecture}

Our model preserves the original video diffusion transformer and its native text/image conditioning pathway. Since both the rendered mesh and depth conditions are represented as RGB videos, they can be encoded by the pretrained video VAE into the same spatiotemporal latent space as the target video. We therefore incorporate the two conditioning signals directly at the input of the diffusion transformer, without introducing additional conditioning encoders or cross-attention branches.

\paragraph{Conditional latent fusion.}
We first encode the rendered mesh and depth conditioning videos using the pretrained causal VAE:
\begin{equation}
z^{mesh}=\mathcal{E}(C^{mesh}),
\qquad
z^{depth}=\mathcal{E}(C^{depth}),
\end{equation}
where $\mathcal{E}$ denotes the VAE encoder.
At each training or sampling step, we concatenate them with the noisy video latent $z_{\tau}$ along the channel dimension:
\begin{equation}
\hat{z}_{\tau}
=
\mathrm{Concat}
\left(
z_{\tau},\,
z^{mesh},\,
z^{depth}
\right),
\end{equation}
For image-to-video generation, the native first-frame conditioning is retained unchanged.

\paragraph{Input adaptation.}
The concatenation increases the input channel dimension of the DiT patch embedding layer. We therefore expand only its input projection to accommodate $\hat{z}_{\tau}$, while keeping the remainder of the pretrained transformer architecture unchanged. The weights corresponding to the original video-latent channels are initialized from the pretrained model, whereas the weights associated with the additional mesh and depth channels are newly initialized.

\paragraph{Flow-matching fine-tuning.}
We fine-tune the model using the same flow-matching formulation as the base video generator. Given a clean target latent $z_0$ and Gaussian noise $\epsilon$, an intermediate latent is constructed along the interpolation path
\begin{equation}
z_{\tau}=(1-\tau)z_0+\tau\epsilon,
\qquad
\tau\sim\mathcal{U}(0,1),
\end{equation}
where $\tau$ denotes the flow-matching time, distinguished from the frame index $t$ used throughout Sec.~\ref{sec:conditioning}, with corresponding target velocity $v_{\tau}=\epsilon-z_0$. Conditioned on the text/image inputs together with $z^{mesh}$ and $z^{depth}$, the network predicts the velocity field
\begin{equation}
\hat{v}_{\tau}
=
u_{\phi}
\left(
\hat{z}_{\tau},\,
\tau,\,
c
\right),
\end{equation}
where $c$ denotes the native text and optional first-frame conditions of the base model, trained with the objective
\begin{equation}
\mathcal{L}_{FM}
=
\mathbb{E}_{z_0,\epsilon,\tau}
\left[
\left\|
\hat{v}_{\tau}-v_{\tau}
\right\|_2^2
\right].
\end{equation}
To preserve the pretrained generation prior while adapting to the additional structural conditions, we freeze the pretrained DiT and adapt it via LoRA, jointly optimized with the expanded patch-embedding layer.

\section{Experiments}

\paragraph{Experimental Setup.}
We construct our training set from Motion-X++~\citep{zhang2025motion}, resulting in 24k training samples. 
We build ReMoVe upon Wan2.2-TI2V-5B~\citep{wan2025} and fine-tune it for conditional video generation. 
The noisy target-video latent is concatenated with the mesh and depth latents along the channel dimension, resulting in a 144-channel input (expanded from the original 48-channel patch embedding). 
Detailed training configurations are provided in Appendix~\ref{sec:add_experiment}.

\subsection{Evaluation Setup}

We evaluate ReMoVe on 152 challenging human-motion videos from Motion-X++ in both the standard video generation setting and a motion-conditioned setting with ground-truth motion input.
We compare against the base generators Wan2.2-TI2V-5B~\citep{wan2025} and CogVideoX-1.5-I2V-5B~\citep{yang2024cogvideox}, the motion-aware method CoMoVi~\citep{zhao2026comovi}, and the motion-conditioned methods EchoMotion~\citep{yang2025echomotion} and MTVCrafter~\citep{ding2025mtvcrafter}.
We assess generation quality with VBench~\citep{huang2023vbench} and VBench-2.0~\citep{zheng2025vbench2}, distributional fidelity with FVD-16 and FID-VID, and reconstruction fidelity with SSIM, LPIPS, and PSNR in the ground-truth-motion setting.
Metric definitions and baseline-specific configurations are provided in Appendix~\ref{sec:add_experiment2}.

\subsection{Comparison with Baselines}

\begin{table*}[t]
\centering
\renewcommand{\arraystretch}{1.2}
\caption{
Quantitative comparison on VBench, the Human Anatomy dimension of VBench-2.0, video-level distribution metrics, and reconstruction fidelity under ground-truth motion.
For the first group of methods, the best and second-best results are highlighted in
\colorbox{orange!45}{darker orange} and
\colorbox{orange!18}{lighter orange}, respectively.
For the GT-motion group, only the best result is highlighted.
}
\label{tab:main_results}
\resizebox{\textwidth}{!}{
\begin{tabular}{lccccccc c cc ccc}
\toprule
Method
& SC & MS & DD & IQ & BC & AQ & TF & HA
& FVD-16$\downarrow$ & FID-VID$\downarrow$
& SSIM$\uparrow$ & LPIPS$\downarrow$ & PSNR$\uparrow$ \\
\midrule

Wan-TI2V-5B
& 91.05 & 98.87 & \hlsecond{66.45} & 67.39 & 92.42 & \hlsecond{49.89} & 97.28 & 90.88
& 91.41 & 38.47
& – & – & – \\

CogVideoX-1.5-I2V-5B
& 92.05 & 98.78 & 9.87 & 65.82 & 93.48 & 47.40 & 97.70 & 75.49
& 75.29 & \hlbest{30.05}
& – & – & – \\

CoMoVi(Wan-5B)
& 88.08 & 96.91 & \hlbest{76.97} & 61.57 & 90.44 & 46.40 & 94.77 & 76.17
& 96.41 & 44.57
& – & – & – \\

\midrule

Ours (mesh)
& 92.21 & 99.30 & 50.66 & \hlsecond{68.82} & 93.88 & 49.47 & 98.80 & 91.54
& \hlbest{63.28} & \hlsecond{30.69}
& – & – & – \\

Ours (depth)
& \hlsecond{93.17} & \hlsecond{99.48} & 42.11 & \hlbest{69.07} & \hlbest{94.60} & 49.36 & \hlsecond{99.25} & \hlsecond{91.87}
& 70.97 & 34.24
& – & – & – \\

Ours (mesh+depth)
& \hlbest{93.26} & \hlbest{99.50} & 40.79 & 68.46 & \hlsecond{94.48} & \hlbest{50.02} & \hlbest{99.27} & \hlbest{92.66}
& \hlsecond{69.87} & 33.00
& – & – & – \\

\midrule

EchoMotion (Wan-5B, GT-motion)
& 85.47 & 94.60 & \hlbest{96.05} & 53.41 & 91.12 & 42.17 & 93.02 & 76.50
& 445.18 & 219.12
& - & - & - \\

MTVCrafter (CogVideoX-5B, GT-motion)
& \hlbest{96.81} & \hlbest{99.71} & 0.0 & \hlbest{70.12} & \hlbest{97.39} & 44.93 & \hlbest{99.80} & \hlbest{94.85}
& 83.27 & 47.28
& 0.7638 & 0.1697 & 19.483 \\

Ours (mesh+depth, GT-motion)
& 94.08 & 99.41 & 11.18 & 68.81 & 94.75 & \hlbest{49.05} & 99.20 & 90.50
& \hlbest{27.71} & \hlbest{16.27}
& \hlbest{0.7984} & \hlbest{0.1214} & \hlbest{22.447} \\

\bottomrule
\end{tabular}}
\end{table*}

\paragraph{Human Motion Video Generation.}
As shown in Tab.~\ref{tab:main_results}, ReMoVe consistently improves the structural and temporal quality of generated human videos over the base video generators and CoMoVi. The full mesh+depth model achieves the best performance on Subject Consistency, Motion Smoothness, Background Consistency, Temporal Flickering, Human Anatomy, and Aesthetic Quality, while maintaining competitive Image Quality. 
It also lowers FVD-16 from 91.41 for Wan-TI2V-5B and 96.41 for CoMoVi to 69.87, while remaining close to CogVideoX-1.5-I2V-5B in FID-VID.

\begin{figure}[t]
\centering
\includegraphics[width=\linewidth]{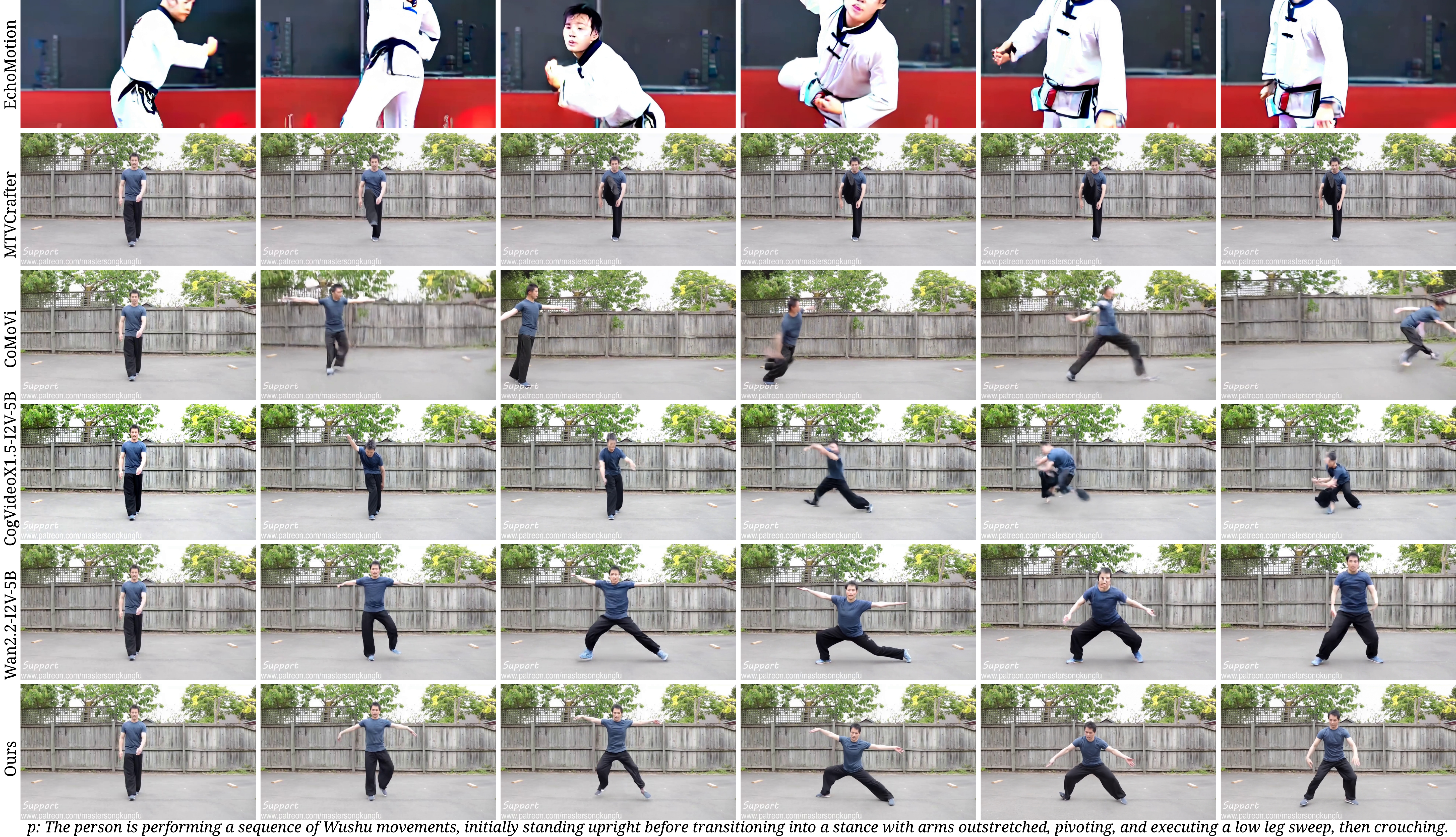}
\caption{Qualitative comparison of human video generation. ReMoVe preserves the original motion prior of Wan while substantially improving human structural stability, appearance consistency, and temporal coherence, outperforming prior methods on challenging human motions.
}
\label{fig:case}
\end{figure}

As shown in Fig.~\ref{fig:case} and the Dynamic Degree results, the baselines exhibit distinct motion behaviors. CogVideoX-1.5-I2V-5B has overly low dynamics, leading to slow and monotonous motion, while Wan-TI2V-5B produces overly high dynamics with abrupt and physically implausible transitions. CoMoVi achieves the highest Dynamic Degree but suffers from degraded Subject Consistency and Human Anatomy. In contrast, ReMoVe maintains a moderate dynamic degree while achieving the best Motion Smoothness and Human Anatomy, benefiting from the combined mesh and depth conditions. Detailed analysis is provided in Appendix~\ref{sec:add_experiment3}.

\paragraph{Motion-Conditioned Video Generation.}
We further evaluate ReMoVe in a motion-conditioned setting, where the ground-truth motion sequence from Motion-X++ is directly provided as input, and compare against EchoMotion and MTVCrafter. 
As shown in Tab.~\ref{tab:main_results} and Fig.~\ref{fig:case}, both baselines exhibit clear limitations. 
EchoMotion achieves a high Dynamic Degree but suffers from severe appearance degradation, poor temporal consistency, and poor reconstruction fidelity. Moreover, it does not support explicit first-frame or reference-image conditioning, so the subject appearance cannot be specified. 
MTVCrafter achieves high consistency-related scores but an extremely low Dynamic Degree of 0.0, indicating that it fails to generate meaningful motion even under large-amplitude ground-truth motion. 
In contrast, ReMoVe achieves the best AQ, FVD-16, FID-VID, SSIM, LPIPS, and PSNR in this motion-conditioned comparison while maintaining stable and non-trivial motion generation, benefiting from explicit mesh-and-depth conditioning. 
Detailed analysis is provided in Appendix~\ref{sec:add_experiment4}.

\subsection{Ablation Studies}
We ablate the rendered mesh and depth conditions to assess their individual contributions. As shown in Tab.~\ref{tab:main_results}, the three variants perform similarly overall because the two signals are complementary but not fully independent. Although introduced to constrain scene geometry, depth also captures the human silhouette and per-frame position, implicitly conveying the subject's motion without the mesh condition.

Mesh-only conditioning substantially improves human structure and temporal stability, achieving a Human Anatomy score of 91.54 and an FVD-16 of 63.28.
However, without explicit scene-level constraints, background and viewpoint changes remain under-constrained. This shows up as a higher Dynamic Degree of 50.66, compared with 42.11 for depth-only and 40.79 for the full model. 
Dynamic Degree captures overall visual dynamics rather than foreground motion alone, so part of this increase likely reflects unintended background or camera variation rather than more meaningful articulated motion.

Depth-only conditioning achieves the best Image Quality and Background Consistency, and still maintains a high Human Anatomy score of 91.87 due to the human geometry implicitly encoded in the depth sequence. However, it provides weaker constraints on articulated body structure and cannot fully correct local deformation or implausible poses inherited from the coarse video.

Combining both signals yields the most balanced performance and the highest Human Anatomy score of 92.66, with the mesh condition constraining pose, body structure, and identity-aware appearance, and the depth condition anchoring scene geometry and background evolution. 
As shown qualitatively in Fig.~\ref{fig:comparison}, mesh-only conditioning tends to preserve human motion but may suffer from background distortion, while depth-only conditioning stabilizes the scene but under-constrains articulated body structure.
Their combination therefore better preserves the coarse video's motion prior while maintaining stable structure, appearance, and scene consistency.

\section{Conclusion}
\label{conclusion}
We presented ReMoVe, which controls human video generation without external motion references by recovering and refining the motion already hypothesized by the base video generator, then translating it into a rendered mesh video and a depth video. These two video-based conditions integrate into a pretrained diffusion transformer through latent concatenation and LoRA adaptation. Experiments show gains in structural stability, appearance consistency, and motion controllability over base generators and existing motion-aware baselines, without requiring ground-truth motion annotations at inference time. More broadly, ReMoVe frames motion control as selective correction of an existing generative prior rather than injection of an external one.

\section*{AI Use Statement}
We acknowledge the use of Claude (Anthropic) and ChatGPT (OpenAI), which provided assistance with language polishing, sentence-level revision, and proofreading during the preparation of this paper. These tools were used to review specific passages for grammatical correctness, clarity, and stylistic naturalness, while preserving the original technical meaning. All suggested edits were reviewed and selectively incorporated by the authors. All research content, experimental design, analyses, and technical interpretations remain entirely the authors' own work.

\bibliography{iclr2027_conference}

@misc{carion2025sam3segmentconcepts,
      title={SAM 3: Segment Anything with Concepts},
      author={Nicolas Carion and Laura Gustafson and Yuan-Ting Hu and Shoubhik Debnath and Ronghang Hu and Didac Suris and Chaitanya Ryali and Kalyan Vasudev Alwala and Haitham Khedr and Andrew Huang and Jie Lei and Tengyu Ma and Baishan Guo and Arpit Kalla and Markus Marks and Joseph Greer and Meng Wang and Peize Sun and Roman Rädle and Triantafyllos Afouras and Effrosyni Mavroudi and Katherine Xu and Tsung-Han Wu and Yu Zhou and Liliane Momeni and Rishi Hazra and Shuangrui Ding and Sagar Vaze and Francois Porcher and Feng Li and Siyuan Li and Aishwarya Kamath and Ho Kei Cheng and Piotr Dollár and Nikhila Ravi and Kate Saenko and Pengchuan Zhang and Christoph Feichtenhofer},
      year={2025},
      eprint={2511.16719},
      archivePrefix={arXiv},
      primaryClass={cs.CV},
      url={https://arxiv.org/abs/2511.16719},
}

@article{depthanything3,
  title={Depth Anything 3: Recovering the visual space from any views},
  author={Haotong Lin and Sili Chen and Jun Hao Liew and Donny Y. Chen and Zhenyu Li and Guang Shi and Jiashi Feng and Bingyi Kang},
  journal={arXiv preprint arXiv:2511.10647},
  year={2025}
}

@inproceedings{shen2024gvhmr,
  title={World-Grounded Human Motion Recovery via Gravity-View Coordinates},
  author={Shen, Zehong and Pi, Huaijin and Xia, Yan and Cen, Zhi and Peng, Sida and Hu, Zechen and Bao, Hujun and Hu, Ruizhen and Zhou, Xiaowei},
  booktitle={SIGGRAPH Asia Conference Proceedings},
  year={2024}
}

@misc{retsinas2025smirk3dfacialexpressions,
      title={SMIRK: 3D Facial Expressions through Analysis-by-Neural-Synthesis}, 
      author={George Retsinas and Panagiotis P. Filntisis and Radek Danecek and Victoria F. Abrevaya and Anastasios Roussos and Timo Bolkart and Petros Maragos},
      year={2025},
      eprint={2404.04104},
      archivePrefix={arXiv},
      primaryClass={cs.CV},
      url={https://arxiv.org/abs/2404.04104}, 
}

@article{flame,
author = {Li, Tianye and Bolkart, Timo and Black, Michael J. and Li, Hao and Romero, Javier},
title = {Learning a model of facial shape and expression from 4D scans},
year = {2017},
issue_date = {December 2017},
publisher = {Association for Computing Machinery},
address = {New York, NY, USA},
volume = {36},
number = {6},
issn = {0730-0301},
url = {https://doi.org/10.1145/3130800.3130813},
doi = {10.1145/3130800.3130813},
journal = {ACM Trans. Graph.},
month = nov,
articleno = {194},
numpages = {17}
}

@article{Kimodo2026,
  title={Kimodo: Scaling Controllable Human Motion Generation},
  author={Rempe, Davis and Petrovich, Mathis and Yuan, Ye and Zhang, Haotian and Peng, Xue Bin and Jiang, Yifeng and Wang, Tingwu and Iqbal, Umar and Minor, David and de Ruyter, Michael and Li, Jiefeng and Tessler, Chen and Lim, Edy and Jeong, Eugene and Wu, Sam and Hassani, Ehsan and Huang, Michael and Yu, Jin-Bey and Chung, Chaeyeon and Song, Lina and Dionne, Olivier and Kautz, Jan and Yuen, Simon and Fidler, Sanja},
  journal={arXiv:2603.15546},
  year={2026}
}

@misc{tevet2022humanmotiondiffusionmodel,
      title={Human Motion Diffusion Model}, 
      author={Guy Tevet and Sigal Raab and Brian Gordon and Yonatan Shafir and Daniel Cohen-Or and Amit H. Bermano},
      year={2022},
      eprint={2209.14916},
      archivePrefix={arXiv},
      primaryClass={cs.CV},
      url={https://arxiv.org/abs/2209.14916}, 
}

@misc{cohan2024flexiblemotioninbetweeningdiffusion,
      title={Flexible Motion In-betweening with Diffusion Models}, 
      author={Setareh Cohan and Guy Tevet and Daniele Reda and Xue Bin Peng and Michiel van de Panne},
      year={2024},
      eprint={2405.11126},
      archivePrefix={arXiv},
      primaryClass={cs.CV},
      url={https://arxiv.org/abs/2405.11126}, 
}

@misc{wen2025hymotion10scalingflow,
      title={HY-Motion 1.0: Scaling Flow Matching Models for Text-To-Motion Generation}, 
      author={Yuxin Wen and Qing Shuai and Di Kang and Jing Li and Cheng Wen and Yue Qian and Ningxin Jiao and Changhai Chen and Weijie Chen and Yiran Wang and Jinkun Guo and Dongyue An and Han Liu and Yanyu Tong and Chao Zhang and Qing Guo and Juan Chen and Qiao Zhang and Youyi Zhang and Zihao Yao and Cheng Zhang and Hong Duan and Xiaoping Wu and Qi Chen and Fei Cheng and Liang Dong and Peng He and Hao Zhang and Jiaxin Lin and Chao Zhang and Zhongyi Fan and Yifan Li and Zhichao Hu and Yuhong Liu and Linus and Jie Jiang and Xiaolong Li and Linchao Bao},
      year={2025},
      eprint={2512.23464},
      archivePrefix={arXiv},
      primaryClass={cs.CV},
      url={https://arxiv.org/abs/2512.23464}, 
}

@misc{rempe2026kimodoscalingcontrollablehuman,
      title={Kimodo: Scaling Controllable Human Motion Generation}, 
      author={Davis Rempe and Mathis Petrovich and Ye Yuan and Haotian Zhang and Xue Bin Peng and Yifeng Jiang and Tingwu Wang and Umar Iqbal and David Minor and Michael de Ruyter and Jiefeng Li and Chen Tessler and Edy Lim and Eugene Jeong and Sam Wu and Ehsan Hassani and Michael Huang and Jin-Bey Yu and Chaeyeon Chung and Lina Song and Olivier Dionne and Jan Kautz and Simon Yuen and Sanja Fidler},
      year={2026},
      eprint={2603.15546},
      archivePrefix={arXiv},
      primaryClass={cs.CV},
      url={https://arxiv.org/abs/2603.15546}, 
}

@misc{zhang2023t2mgptgeneratinghumanmotion,
      title={T2M-GPT: Generating Human Motion from Textual Descriptions with Discrete Representations}, 
      author={Jianrong Zhang and Yangsong Zhang and Xiaodong Cun and Shaoli Huang and Yong Zhang and Hongwei Zhao and Hongtao Lu and Xi Shen},
      year={2023},
      eprint={2301.06052},
      archivePrefix={arXiv},
      primaryClass={cs.CV},
      url={https://arxiv.org/abs/2301.06052}, 
}

@misc{chefer2025videojamjointappearancemotionrepresentations,
      title={VideoJAM: Joint Appearance-Motion Representations for Enhanced Motion Generation in Video Models}, 
      author={Hila Chefer and Uriel Singer and Amit Zohar and Yuval Kirstain and Adam Polyak and Yaniv Taigman and Lior Wolf and Shelly Sheynin},
      year={2025},
      eprint={2502.02492},
      archivePrefix={arXiv},
      primaryClass={cs.CV},
      url={https://arxiv.org/abs/2502.02492}, 
}

@article{zhang2025fantasyid,
  title={Fantasyid: Face knowledge enhanced id-preserving video generation},
  author={Zhang, Yunpeng and Wang, Qiang and Jiang, Fan and Fan, Yaqi and Xu, Mu and Qi, Yonggang},
  journal={arXiv preprint arXiv:2502.13995},
  year={2025}
}

@inproceedings{hu2025animate,
  title={Animate anyone 2: High-fidelity character image animation with environment affordance},
  author={Hu, Li and Wang, Guangyuan and Shen, Zhen and Gao, Xin and Meng, Dechao and Zhuo, Lian and Zhang, Peng and Zhang, Bang and Bo, Liefeng},
  booktitle={2025 IEEE/CVF International Conference on Computer Vision (ICCV)},
  pages={10207--10217},
  year={2025},
  organization={IEEE}
}

@inproceedings{zhu2024champ,
      title={Champ: Controllable and Consistent Human Image Animation with 3D Parametric Guidance},
      author={Shenhao Zhu and Junming Leo Chen and Zuozhuo Dai and Yinghui Xu and Xun Cao and Yao Yao and Hao Zhu and Siyu Zhu},
      booktitle={European Conference on Computer Vision (ECCV)},
      year={2024}
}

@article{wang2025humandreamer,
        title={HumanDreamer: Generating Controllable Human-Motion Videos via Decoupled Generation}, 
        author={Boyuan Wang and Xiaofeng Wang and Chaojun Ni and Guosheng Zhao and Zhiqin Yang and Zheng Zhu and Muyang Zhang and Yukun Zhou and Xinze Chen and Guan Huang and Lihong Liu and Xingang Wang},
        journal={arXiv preprint arXiv:2503.24026},
        year={2025}
}

@article{yang2025echomotion,
  title={EchoMotion: Unified Human Video and Motion Generation via Dual-Modality Diffusion Transformer},
  author={Yang, Yuxiao and Sheng, Hualian and Cai, Sijia and Lin, Jing and Wang, Jiahao and Deng, Bing and Lu, Junzhe and Wang, Haoqian and Ye, Jieping},
  journal={arXiv preprint arXiv:2512.18814},
  year={2025}
}

@InProceedings{zhao2026comovi,
  title   = {CoMoVi: Co-Generation of 3D Human Motions and Realistic Videos},
  author  = {Zhao, Chengfeng and Shu, Jiazhi and Zhao, Yubo and Huang, Tianyu and Lu, Jiahao and Gu, Zekai and Ren, Chengwei and Dou, Zhiyang and Shuai, Qing and Liu, Yuan},
  journal = {arXiv preprint arXiv:2601.10632},
  year    = {2026}
}

@misc{taghipour2026controllablecomplexhumanmotion,
      title={Controllable Complex Human Motion Video Generation via Text-to-Skeleton Cascades}, 
      author={Ashkan Taghipour and Morteza Ghahremani and Zinuo Li and Hamid Laga and Farid Boussaid and Mohammed Bennamoun},
      year={2026},
      eprint={2603.08028},
      archivePrefix={arXiv},
      primaryClass={cs.CV},
      url={https://arxiv.org/abs/2603.08028}, 
}

@misc{xu2025hunyuanportraitimplicitconditioncontrol,
      title={HunyuanPortrait: Implicit Condition Control for Enhanced Portrait Animation}, 
      author={Zunnan Xu and Zhentao Yu and Zixiang Zhou and Jun Zhou and Xiaoyu Jin and Fa-Ting Hong and Xiaozhong Ji and Junwei Zhu and Chengfei Cai and Shiyu Tang and Qin Lin and Xiu Li and Qinglin Lu},
      year={2025},
      eprint={2503.18860},
      archivePrefix={arXiv},
      primaryClass={cs.CV},
      url={https://arxiv.org/abs/2503.18860}, 
}

@article{guo2026wildactor,
  title={WildActor: Unconstrained Identity-Preserving Video Generation},
  author={Guo, Qin and Yang, Tianyu and He, Xuanhua and Shen, Fei and Zhang, Yong and Kang, Zhuoliang and Wei, Xiaoming and Xu, Dan},
  journal={arXiv preprint arXiv:2603.00586},
  year={2026}
}

@article{cao2025uni3c,
        title={Uni3C: Unifying Precisely 3D-Enhanced Camera and Human Motion Controls for Video Generation},
        author={Cao, Chenjie and Zhou, Jingkai and Li, shikai and Liang, Jingyun and Yu, Chaohui and Wang, Fan and Xue, Xiangyang and Fu, Yanwei},
        journal={arXiv preprint arXiv:2504.14899},
        year={2025}}

@misc{liu2026revisionrefiningvideodiffusion,
      title={ReVision: Refining Video Diffusion with Explicit 3D Motion Modeling}, 
      author={Qihao Liu and Ju He and Qihang Yu and Liang-Chieh Chen and Alan Yuille},
      year={2026},
      eprint={2504.21855},
      archivePrefix={arXiv},
      primaryClass={cs.CV},
      url={https://arxiv.org/abs/2504.21855}, 
}

@article{wan2025,
      title={Wan: Open and Advanced Large-Scale Video Generative Models}, 
      author={Team Wan and Ang Wang and Baole Ai and Bin Wen and Chaojie Mao and Chen-Wei Xie and Di Chen and Feiwu Yu and Haiming Zhao and Jianxiao Yang and Jianyuan Zeng and Jiayu Wang and Jingfeng Zhang and Jingren Zhou and Jinkai Wang and Jixuan Chen and Kai Zhu and Kang Zhao and Keyu Yan and Lianghua Huang and Mengyang Feng and Ningyi Zhang and Pandeng Li and Pingyu Wu and Ruihang Chu and Ruili Feng and Shiwei Zhang and Siyang Sun and Tao Fang and Tianxing Wang and Tianyi Gui and Tingyu Weng and Tong Shen and Wei Lin and Wei Wang and Wei Wang and Wenmeng Zhou and Wente Wang and Wenting Shen and Wenyuan Yu and Xianzhong Shi and Xiaoming Huang and Xin Xu and Yan Kou and Yangyu Lv and Yifei Li and Yijing Liu and Yiming Wang and Yingya Zhang and Yitong Huang and Yong Li and You Wu and Yu Liu and Yulin Pan and Yun Zheng and Yuntao Hong and Yupeng Shi and Yutong Feng and Zeyinzi Jiang and Zhen Han and Zhi-Fan Wu and Ziyu Liu},
      journal = {arXiv preprint arXiv:2503.20314},
      year={2025}
}

@article{yang2024cogvideox,
  title={CogVideoX: Text-to-Video Diffusion Models with An Expert Transformer},
  author={Yang, Zhuoyi and Teng, Jiayan and Zheng, Wendi and Ding, Ming and Huang, Shiyu and Xu, Jiazheng and Yang, Yuanming and Hong, Wenyi and Zhang, Xiaohan and Feng, Guanyu and others},
  journal={arXiv preprint arXiv:2408.06072},
  year={2024}
}

@InProceedings{huang2023vbench,
  title={{VBench}: Comprehensive Benchmark Suite for Video Generative Models},
  author={Huang, Ziqi and He, Yinan and Yu, Jiashuo and Zhang, Fan and Si, Chenyang and Jiang, Yuming and Zhang, Yuanhan and Wu, Tianxing and Jin, Qingyang and Chanpaisit, Nattapol and Wang, Yaohui and Chen, Xinyuan and Wang, Limin and Lin, Dahua and Qiao, Yu and Liu, Ziwei},
  booktitle={Proceedings of the IEEE/CVF Conference on Computer Vision and Pattern Recognition},
  year={2024}
}

@article{zheng2025vbench2,
      title={{VBench-2.0}: Advancing Video Generation Benchmark Suite for Intrinsic Faithfulness},
      author={Zheng, Dian and Huang, Ziqi and Liu, Hongbo and Zou, Kai and He, Yinan and Zhang, Fan and Zhang, Yuanhan and He, Jingwen and Zheng, Wei-Shi and Qiao, Yu and Liu, Ziwei},
      journal={arXiv preprint arXiv:2503.21755},
      year={2025}
}

@article{zhang2025motion,
  title={Motion-X++: A Large-Scale Multimodal 3D Whole-body Human Motion Dataset},
  author={Zhang, Yuhong and Lin, Jing and Zeng, Ailing and Wu, Guanlin and Lu, Shunlin and Fu, Yurong and Cai, Yuanhao and Zhang, Ruimao and Wang, Haoqian and Zhang, Lei},
  journal={arXiv preprint arXiv:2501.05098},
  year={2025}
}

@InProceedings{Yu_2025_ICCV,
    author    = {Yu, Mark and Hu, Wenbo and Xing, Jinbo and Shan, Ying},
    title     = {TrajectoryCrafter: Redirecting Camera Trajectory for Monocular Videos via Diffusion Models},
    booktitle = {Proceedings of the IEEE/CVF International Conference on Computer Vision (ICCV)},
    month     = {October},
    year      = {2025},
    pages     = {100-111}
}

@misc{wang2026epicefficientvideocamera,
      title={EPiC: Efficient Video Camera Control Learning with Precise Anchor-Video Guidance}, 
      author={Zun Wang and Jaemin Cho and Jialu Li and Han Lin and Jaehong Yoon and Yue Zhang and Mohit Bansal},
      year={2026},
      eprint={2505.21876},
      archivePrefix={arXiv},
      primaryClass={cs.CV},
      url={https://arxiv.org/abs/2505.21876}, 
}

@article{li2025realcam,
    title={RealCam-I2V: Real-World Image-to-Video Generation with Interactive Complex Camera Control}, 
    author={Li, Teng and Zheng, Guangcong and Jiang, Rui and Zhan, Shuigen and Wu, Tao and Lu, Yehao and Lin, Yining and Li, Xi},
    journal={arXiv preprint arXiv:2502.10059},
    year={2025},
}

@misc{wang2025vggtvisualgeometrygrounded,
      title={VGGT: Visual Geometry Grounded Transformer}, 
      author={Jianyuan Wang and Minghao Chen and Nikita Karaev and Andrea Vedaldi and Christian Rupprecht and David Novotny},
      year={2025},
      eprint={2503.11651},
      archivePrefix={arXiv},
      primaryClass={cs.CV},
      url={https://arxiv.org/abs/2503.11651}, 
}

@article{ding2025mtvcrafter,
  title={MTVCrafter: 4D Motion Tokenization for Open-World Human Image Animation},
  author={Ding, Yanbo and Hu, Xirui and Guo, Zhizhi and Zhang, Chi and Wang, Yali},
  journal={arXiv preprint arXiv:2505.10238},
  year={2025}
}

@misc{he2025cameractrliidynamicscene,
      title={CameraCtrl II: Dynamic Scene Exploration via Camera-controlled Video Diffusion Models}, 
      author={Hao He and Ceyuan Yang and Shanchuan Lin and Yinghao Xu and Meng Wei and Liangke Gui and Qi Zhao and Gordon Wetzstein and Lu Jiang and Hongsheng Li},
      year={2025},
      eprint={2503.10592},
      archivePrefix={arXiv},
      primaryClass={cs.CV},
      url={https://arxiv.org/abs/2503.10592}, 
}

@article{kong2024hunyuanvideo,
  title={Hunyuanvideo: A systematic framework for large video generative models},
  author={Kong, Weijie and Tian, Qi and Zhang, Zijian and Min, Rox and Dai, Zuozhuo and Zhou, Jin and Xiong, Jiangfeng and Li, Xin and Wu, Bo and Zhang, Jianwei and others},
  journal={arXiv preprint arXiv:2412.03603},
  year={2024}
}

@misc{fang2026motioncharacterfinegrainedmotioncontrollable,
      title={MotionCharacter: Fine-Grained Motion Controllable Human Video Generation}, 
      author={Haopeng Fang and Di Qiu and Binjie Mao and He Tang},
      year={2026},
      eprint={2411.18281},
      archivePrefix={arXiv},
      primaryClass={cs.CV},
      url={https://arxiv.org/abs/2411.18281}, 
}

@inproceedings{tu2025stableanimator,
  title={Stableanimator: High-quality identity-preserving human image animation},
  author={Tu, Shuyuan and Xing, Zhen and Han, Xintong and Cheng, Zhi-Qi and Dai, Qi and Luo, Chong and Wu, Zuxuan},
  booktitle={Proceedings of the Computer Vision and Pattern Recognition Conference},
  pages={21096--21106},
  year={2025}
}
\bibliographystyle{iclr2027_conference}

\clearpage
\appendix
\section{Appendix}

\subsection{Temporal Stage-Assignment Conflict}
\label{sec:appendix_temporal_conflict}

\begin{figure}[!htbp]
\centering

\begin{minipage}[t]{0.49\linewidth}
    \centering
    \includegraphics[width=\linewidth]
    {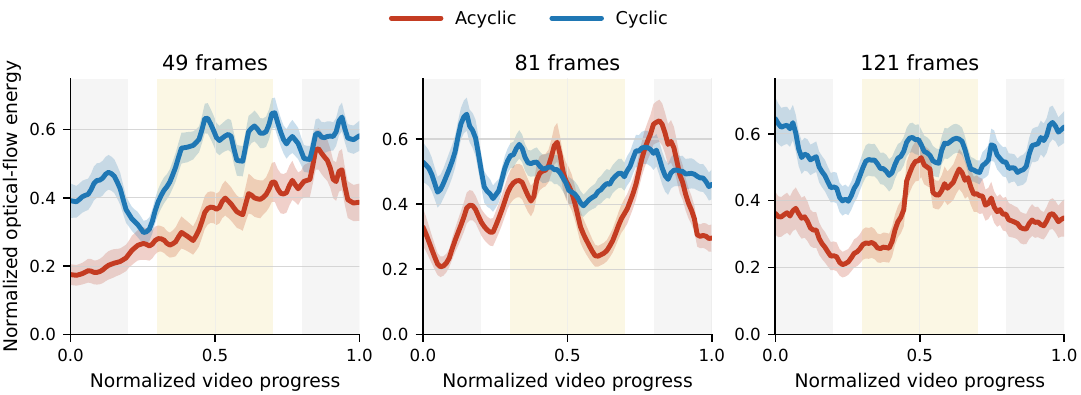}
    \par\vspace{1mm}
    {\small (a) Optical-flow motion energy\par}
\end{minipage}
\hfill
\begin{minipage}[t]{0.49\linewidth}
    \centering
    \includegraphics[width=\linewidth]
    {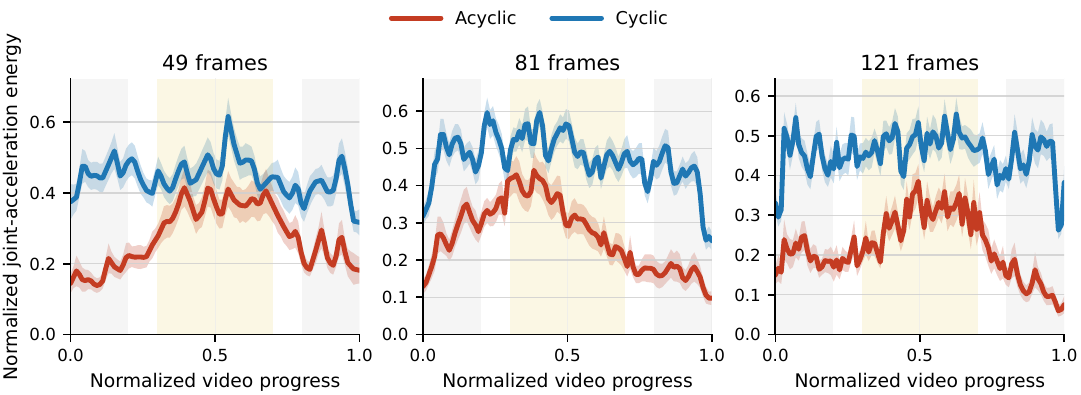}
    \par\vspace{1mm}
    {\small (b) GVHMR joint acceleration\par}
\end{minipage}

\vspace{3mm}

{\small
\setlength{\tabcolsep}{6pt}
\renewcommand{\arraystretch}{1.08}
\begin{tabular}{@{}llcccc@{}}
\toprule
Frames & Type & CV & M/E & P/M & Jerk \\
\midrule
49  & Acy. & 0.918 & 2.50 & 4.51 & 4.88 \\
    & Cyc. & 0.514 & 1.32 & 2.85 & 3.09 \\
\addlinespace[2pt]
81  & Acy. & 0.934 & 2.24 & 4.91 & 5.77 \\
    & Cyc. & 0.423 & 1.18 & 2.48 & 2.77 \\
\addlinespace[2pt]
121 & Acy. & 1.007 & 3.25 & 5.77 & 7.11 \\
    & Cyc. & 0.448 & 1.13 & 2.83 & 3.68 \\
\bottomrule
\end{tabular}
\par}

\caption{
Temporal motion analysis for acyclic and cyclic human actions
under three frame budgets.
Top: optical-flow motion energy and GVHMR joint-acceleration
energy over normalized video progress.
Bottom: GVHMR joint-acceleration statistics.
CV denotes the coefficient of variation of joint acceleration.
M/E denotes the middle-to-edge acceleration ratio.
P/M denotes the peak-to-mean acceleration ratio.
Jerk denotes the peak-to-mean jerk ratio.
Across the tested budgets, acyclic actions exhibit greater
temporal concentration of motion than cyclic actions.
}
\label{fig:temporal_motion_analysis}
\end{figure}

We empirically analyze how complex goal-directed actions and repetitive actions distribute motion under a fixed temporal budget.
We compare two categories of generated human videos: \textit{goal-directed acyclic} actions, such as backflips, aerial kicks, and single-turn martial-arts motions, and \textit{repetitive cyclic} actions, such as repeated footwork, alternating kicks, and shadowboxing-like motions.
For each acyclic prompt, we construct a paired cyclic prompt using the same first frame and scene context, so that the comparison primarily reflects the temporal structure of the requested motion rather than differences in appearance or viewpoint.

\paragraph{Setup.}
We evaluate 180 generated videos in total.
The dataset includes two temporal budgets: 49 frames, 81 frames and 121 frames.
For each budget, we use 30 acyclic videos and 30 paired cyclic videos.
All paired videos are generated from identical first-frame conditions.

\paragraph{Optical-flow motion profile.}
We first measure the video-level motion distribution using dense optical flow.
For each pair of consecutive frames, we compute the average optical-flow magnitude as the frame-wise motion energy.
Each video's motion-energy curve is normalized by its maximum value and then averaged within each group.
As shown in Fig.~\ref{fig:temporal_motion_analysis}, cyclic motions tend to maintain sustained motion for most of the video, apart from the initial action onset, whereas acyclic motions exhibit more localized peaks, followed by a sharp drop near the end.
Consistent with our manual observation, 28 of 30 acyclic videos in the 121-frame setting, 24 of 30 acyclic videos in the 81-frame setting and 28 of 30 acyclic videos in the 49-frame setting exhibit this pattern.
This suggests that repetitive actions can be continued or truncated more smoothly, whereas goal-directed actions concentrate motion around specific action stages, which is also where abrupt motion changes are most likely to occur.

\paragraph{Joint-acceleration analysis.}
To measure human motion changes more directly, we estimate 3D human motion using GVHMR and compute joint acceleration from the recovered SMPL joints.
Let $J_t \in \mathbb{R}^{K \times 3}$ denote the 3D joint positions at frame $t$.
We remove global root translation by centering all joints at the pelvis, and compute acceleration as
\begin{equation}
A_t = J_{t+1} - 2J_t + J_{t-1}.
\end{equation}
The frame-wise acceleration energy is then
\begin{equation}
E_t = \frac{1}{K}\sum_{k=1}^{K} \|A_{t,k}\|_2 .
\end{equation}
We also compute jerk, the third-order temporal difference, to capture sharper changes in joint motion.

The quantitative results consistently support our observation.
Across the 49-frame, 81-frame, and 121-frame settings, acyclic actions exhibit substantially higher acceleration CV, middle-to-edge ratio, peak-to-mean acceleration ratio, and jerk peak-to-mean ratio than cyclic actions.
For example, in the 81-frame setting, acyclic actions have an acceleration CV of $0.934$ and a peak-to-mean acceleration ratio of $4.91$, compared to $0.423$ and $2.48$ for cyclic actions.
This distinction persists at 121 frames, where the corresponding values are $1.007$ and $5.77$ for acyclic actions, versus $0.448$ and $2.83$ for cyclic actions.
As shown in Fig.~\ref{fig:temporal_motion_analysis}, the joint-acceleration profiles of acyclic actions consistently exhibit stronger motion changes during the intermediate portion of the video and lower activity toward both ends, whereas cyclic actions maintain more sustained activity.
The optical-flow profiles likewise reveal localized periods of motion, although their timing varies across settings.
Notably, the 121-frame acyclic profile exhibits a more sustained low-acceleration tail.
Consistent with our manual observations of fewer abrupt changes near the end of these videos, this suggests that the longer budget provides more room for the action to reach its final state and settle.
The higher whole-video peak-to-mean ratios are compatible with this quieter ending, since a longer low-motion tail can reduce the temporal mean.

Cyclic actions remain plausible when their local motion patterns are repeated, continued, or truncated.
In contrast, acyclic actions must pass through a specific sequence of stages, such as preparation, takeoff, rotation, and landing.
When the model is asked to synthesize such an action under a fixed temporal budget without explicit stage-wise supervision, it may compress or skip intermediate transitions, resulting in abrupt motion changes and unstable intermediate poses.
ReMoVe addresses this conflict by decoupling the motion-refinement horizon from the coarse video's original frame budget.
The motion refiner synthesizes a motion sequence within a longer generation window, after which trailing near-static frames are removed.
The resulting mesh video provides explicit guidance for synthesizing complex goal-directed actions with coherent stage transitions and stable appearance.

\subsection{Camera-Controlled Video Generation}

\paragraph{Experimental Setup.}
We evaluate camera controllability on 30 martial-arts action clips sampled from Motion-X++, covering a diverse range of human motion dynamics. For each clip, we define five target camera trajectories: \textit{dolly-in} (0.5\,m forward translation), \textit{truck-left} (0.5\,m lateral translation), \textit{pedestal-up} (0.3\,m vertical translation), \textit{pan-left} ($20^\circ$ in-place yaw rotation), and \textit{orbit} ($40^\circ$ rotation about the vertical axis around a pivot point located 2\,m in front of the camera, coupling translation and rotation). This results in 152 clip--trajectory test instances for each method. Due to their native generation configurations, TrajectoryCrafter~\citep{Yu_2025_ICCV} and EPiC~\citep{wang2026epicefficientvideocamera} generate 49-frame videos, whereas RealCam-I2V~\citep{li2025realcam} and ReMoVe generate 81-frame videos.

\begin{table}[htbp]
\centering
\caption{
Camera-control evaluation on simple translational trajectories, including
\textit{dolly-in}, \textit{truck-left}, and \textit{pedestal-up}.
Lower is better for all metrics.
}
\label{tab:camera_translation}
\begin{tabular}{lcccccc}
\toprule
Method
& ATE$\downarrow$
& RPE$\downarrow$
& RRE ($^\circ$)$\downarrow$
& RotErr ($^\circ$)$\downarrow$
& TransErr$\downarrow$
& CamMC$\downarrow$ \\
\midrule
RealCam-I2V
& 0.0509
& \textbf{0.0279}
& 1.1614
& 0.6732
& 0.1466
& 0.1511 \\

EPIC
& 0.0475
& 0.0300
& \textbf{0.2631}
& \textbf{0.4585}
& 0.1241
& 0.1254 \\

TrajCrafter
& \textbf{0.0465}
& 0.0301
& 0.2714
& 0.4786
& \textbf{0.1119}
& \textbf{0.1133} \\

\midrule
Ours
& 0.0584
& 0.0341
& 0.3005
& 0.4712
& 0.2123
& 0.2132 \\
\bottomrule
\end{tabular}
\end{table}

\begin{table}[htbp]
\centering
\caption{
Camera-control evaluation on challenging rotational and coupled trajectories,
including \textit{pan-left} and \textit{orbit}.
RRE and RotErr are averaged over both trajectories, while
translation-dependent metrics (ATE, RPE, TransErr, and CamMC) are evaluated
on \textit{orbit} only, since \textit{pan-left} contains no ground-truth camera translation.
Lower is better for all metrics.
}
\label{tab:camera_rotation2}
\begin{tabular}{lcccccc}
\toprule
Method
& ATE$\downarrow$
& RPE$\downarrow$
& RRE ($^\circ$)$\downarrow$
& RotErr ($^\circ$)$\downarrow$
& TransErr$\downarrow$
& CamMC$\downarrow$ \\
\midrule
RealCam-I2V
& 0.0665
& 0.0384
& 1.7816
& 3.5185
& 0.0494
& 0.1092 \\

\midrule
Ours
& \textbf{0.0521}
& \textbf{0.0322}
& \textbf{1.3440}
& \textbf{2.7766}
& \textbf{0.0361}
& \textbf{0.0839} \\
\bottomrule
\end{tabular}
\end{table}

\paragraph{Baselines.}
We compare against three representative camera-controlled video generation methods: RealCam-I2V~\citep{li2025realcam}, TrajectoryCrafter~\citep{Yu_2025_ICCV}, and EPiC~\citep{wang2026epicefficientvideocamera}. RealCam-I2V~\citep{li2025realcam} directly conditions video generation on a sequence of metric-scale camera poses encoded as per-frame Pl\"ucker coordinates. TrajectoryCrafter~\citep{Yu_2025_ICCV} lifts a source video into a dynamic point cloud using monocular depth estimation, renders it along the prescribed camera trajectory, and jointly conditions a dual-stream video diffusion model on the resulting point-cloud renders and the source video. For EPiC~\citep{wang2026epicefficientvideocamera}, we adopt its video-to-video camera-control setting, where a source video is used to construct dynamic point clouds and the corresponding anchor video for generation.

Notably, ReMoVe and RealCam-I2V~\citep{li2025realcam} operate in a pure image-to-video setting and require only the first frame as visual input, whereas TrajectoryCrafter~\citep{Yu_2025_ICCV} and EPiC~\citep{wang2026epicefficientvideocamera} additionally take the original video as input in our evaluation. For the latter two methods, we use the corresponding original videos provided by Motion-X++ as their source videos. For ReMoVe, the prescribed camera trajectory is consistently applied to both the rendered mesh and depth conditions, providing geometrically aligned human-structure and scene-level guidance throughout generation.

\begin{figure}[htbp]
\centering
\includegraphics[width=\linewidth]{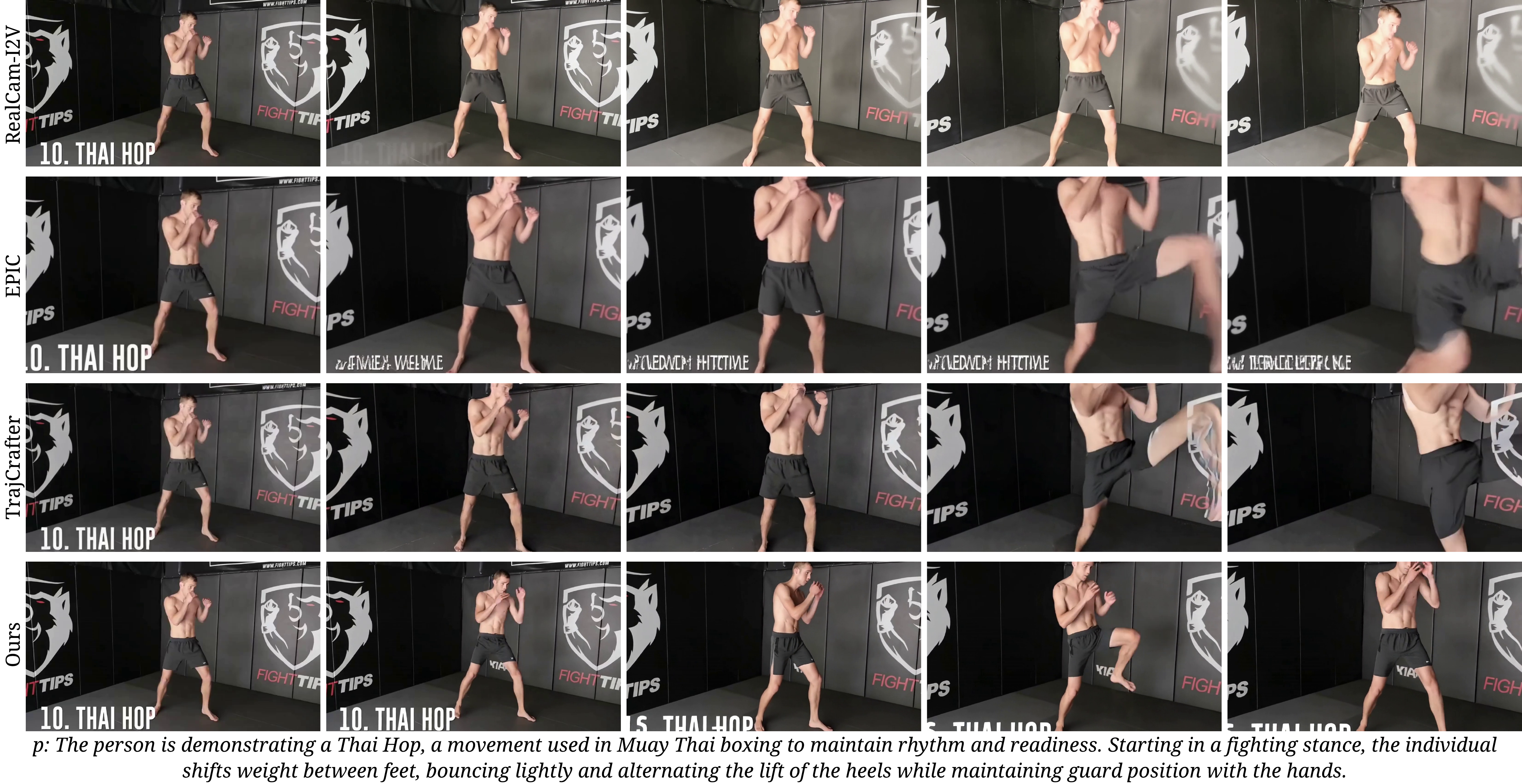}
\caption{Qualitative comparison of camera-controlled video generation under the \textit{dolly-in} trajectory. We compare ReMoVe with RealCam-I2V, TrajectoryCrafter, and EPiC on the same input sample and prescribed camera motion.}
\label{fig:camera_case}
\end{figure}

\paragraph{Evaluation Protocol and Metrics.}
For each generated video, we recover its camera trajectory using VGGT~\citep{wang2025vggtvisualgeometrygrounded} instead of classical structure-from-motion pipelines, which we found unreliable on diffusion-generated videos due to imperfect cross-frame geometric consistency. We uniformly sample 10 frames spanning the full video duration for camera pose estimation. Since monocular camera reconstruction is subject to an unknown global coordinate frame and scale, we align the estimated trajectory to the target trajectory before evaluation. 
Specifically, we estimate the global rotation by Kabsch averaging over camera orientations, determine the scale from the ratio of trajectory path lengths for trajectories with non-zero translation, and recover the global translation by least-squares alignment. 
For pure-rotation trajectories, the translation scale is undefined and is therefore not used for evaluation. We report six lower-is-better camera-control metrics: Absolute Trajectory Error (ATE) and translational Relative Pose Error (RPE) for camera-position accuracy, Relative Rotation Error (RRE) for frame-to-frame rotational accuracy, and RotErr, TransErr, and CamMC following prior camera-control evaluation protocols. RotErr measures the mean geodesic rotation discrepancy, TransErr measures the normalized translation discrepancy, and CamMC jointly evaluates rotation and translation using camera-to-world matrices. For the pure-rotation \textit{pan-left} trajectory, whose target camera center remains fixed, translation-dependent metrics are ill-defined under scene-scale normalization. Therefore, RRE and RotErr are evaluated over both \textit{pan-left} and \textit{orbit}, while ATE, RPE, TransErr, and CamMC are evaluated on \textit{orbit} only.

\paragraph{Results.}

As shown in Tab.~\ref{tab:camera_translation}, all four methods achieve comparable camera-control accuracy on the three simple translational trajectories. 
Although ReMoVe does not obtain the best score on these metrics, the differences remain relatively small. 
Notably, this comparison is less favorable to ReMoVe in terms of available visual information: ReMoVe and RealCam-I2V operate in an image-to-video setting using only the first frame, whereas TrajectoryCrafter and EPiC additionally have access to the complete source video in our evaluation.
Nevertheless, ReMoVe remains competitive with these dedicated camera-control methods, indicating that the camera motion encoded through our geometrically aligned mesh and depth conditions is sufficient to provide effective control for simple translations.

The advantage of ReMoVe becomes more evident on the challenging rotational and coupled trajectories in Tab.~\ref{tab:camera_rotation2}. 
TrajectoryCrafter and EPiC do not provide an equivalent per-frame control interface for these presets under our evaluation setup, so we instead compare against RealCam-I2V, which explicitly supports full camera-pose sequences.
ReMoVe consistently outperforms RealCam-I2V across all six metrics, reducing errors in both camera rotation and translation. 
In particular, the gains on \textit{pan-left} and \textit{orbit} demonstrate that applying the prescribed trajectory consistently to both the rendered mesh and depth conditions enables the generated video to better follow rotational and coupled camera motion. 
These results suggest that the proposed conditioning design stabilizes human motion and scene geometry while also providing effective camera controllability, without requiring a dedicated camera-control module.
\subsection{Implementation Details of the Motion Refiner}
\label{sec:appendix_motion_refiner}

\paragraph{Physical-plausibility metrics.}
For each recovered pose $\hat\theta_t$ we forward-kinematically compute the body joint positions $J_t\in\mathbb{R}^{22\times3}$ of the SMPL-X kinematic tree, excluding hands and face, using the shape-corrected T-pose template. Let $y_t^{L}$ and $y_t^{R}$ denote the height of the left and right foot joint at frame $t$, each taken as the minimum of the ankle and toe joint heights on that side. We estimate a per-sequence ground plane height as the second percentile of the foot-height signal,
\begin{equation}
y_{\text{gnd}} = \operatorname{Percentile}_{2\%}\!\big(\{\min(y_t^{L},y_t^{R})\}_{t=1}^{T}\big),
\end{equation}
and mark a foot as in contact with the ground when its height falls within a tolerance of $0.05\,\text{m}$ of $y_{\text{gnd}}$. We compute the following physical-plausibility signals at every frame $t$. The ground penetration $gp_t$ is the amount by which the lower foot falls below the ground plane, $gp_t = \max(0,\, y_{\text{gnd}} - \min(y_t^{L}, y_t^{R}))$. The foot-sliding velocity $fs_t$ is the horizontal displacement of a foot joint between consecutive frames, evaluated only while that foot remains in contact across both frames. Finally, the root velocity, acceleration, and jerk, $\dot r_t$, $\ddot r_t$, and $\dddot r_t$, are obtained by finite differences of the root translation sampled at $30\,\text{fps}$.

For each sequence, we summarize $gp_t$ by its mean and maximum over all frames, and $fs_t$ by its median over contact frames, denoted $\overline{fs}_c$. We use the median rather than the mean to suppress noisy landing-frame outliers. We further summarize $\ddot r_t$, $\dddot r_t$, and $\|\dot r_t\|$ by their maxima over the sequence.

\paragraph{Reliability tiering.}
Each sequence is assigned a reliability tier $\tau \in \{\texttt{low}, \texttt{medium}, \texttt{high}\}$ based on these sequence-level statistics. A sequence is labeled \texttt{low} whenever it shows any sign of severe artifacts: its median contact-phase foot-sliding velocity reaches $1.0\,\text{m/s}$, its peak foot-sliding velocity reaches $3.0\,\text{m/s}$, its peak root acceleration reaches $70\,\text{m/s}^2$, its peak root jerk reaches $3000\,\text{m/s}^3$, or its peak root speed reaches $8.0\,\text{m/s}$.
Among the remaining sequences, a sequence is labeled \texttt{high} when its median contact-phase foot-sliding velocity stays below $0.4\,\text{m/s}$ and its peak root acceleration and peak root jerk stay below $35\,\text{m/s}^2$ and $2000\,\text{m/s}^3$, respectively. All other sequences are labeled \texttt{medium}.
On the 152 test videos, this procedure assigns 51 sequences (33.6\%) to \texttt{low}, 54 (35.5\%) to \texttt{medium}, and 47 (30.9\%) to \texttt{high}.

\paragraph{Anchor frame selection $\mathcal{K}$.}
Independent of the sequence-level tier, we additionally flag individual frames as unreliable under a tighter, frame-level set of thresholds. A frame is considered unreliable if its ground penetration exceeds $0.02\,\text{m}$, its foot-sliding velocity exceeds $1.5\,\text{m/s}$, its root acceleration exceeds $15\,\text{m/s}^2$, its root jerk exceeds $2000\,\text{m/s}^3$, or its root speed exceeds $5\,\text{m/s}$.

The retained frame index set $\mathcal{K}$ is then constructed according to the sequence tier. For \texttt{high} sequences, all frames are kept as anchors, $\mathcal{K}=\{1,\dots,T\}$. For \texttt{low} sequences, $\mathcal{K}$ retains only the first five frames, providing a coarse pose and shape prior while discarding the unreliable remainder.
For \texttt{medium} sequences, $\mathcal{K}$ is formed by merging the reliable frames, bridging isolated unreliable gaps of at most two frames and discarding reliable runs shorter than five frames. This better preserves locally trustworthy motion segments than a fixed-length prefix.
The first frame is always included as an anchor in both the \texttt{low} and \texttt{medium} cases.
Each anchor frame contributes its local joint rotations, along with a canonicalized root position and trajectory, to the constraint set passed to $\mathcal{R}$.
Here, canonicalization means the heading and initial horizontal position are aligned to a fixed reference.

\paragraph{Keyframe-guided synthesis and trimming.}
For \texttt{low} and \texttt{medium} sequences, the sparse anchor set does not by itself span the full target duration, so we query $\mathcal{R}$ with a longer generation horizon of five seconds at $30\,\text{fps}$, giving the synthesis model sufficient temporal context to produce a coherent full-length trajectory conditioned on $\{\hat\theta_t\}_{t\in\mathcal{K}}$ and $\mathbf{y}$. The output is then trimmed by detecting the last frame whose joint velocity exceeds the sequence's mean joint velocity and truncating any trailing near-static frames beyond it, yielding $\Theta^{ref}$ at the target length $T$. For \texttt{high} sequences, $\mathcal{K}$ already spans the entire clip, so $\mathcal{R}$ is queried directly at the original duration $T/30\,\text{fps}$, and no trimming step is required.

\begin{figure}[t]
\centering
\includegraphics[width=\linewidth]{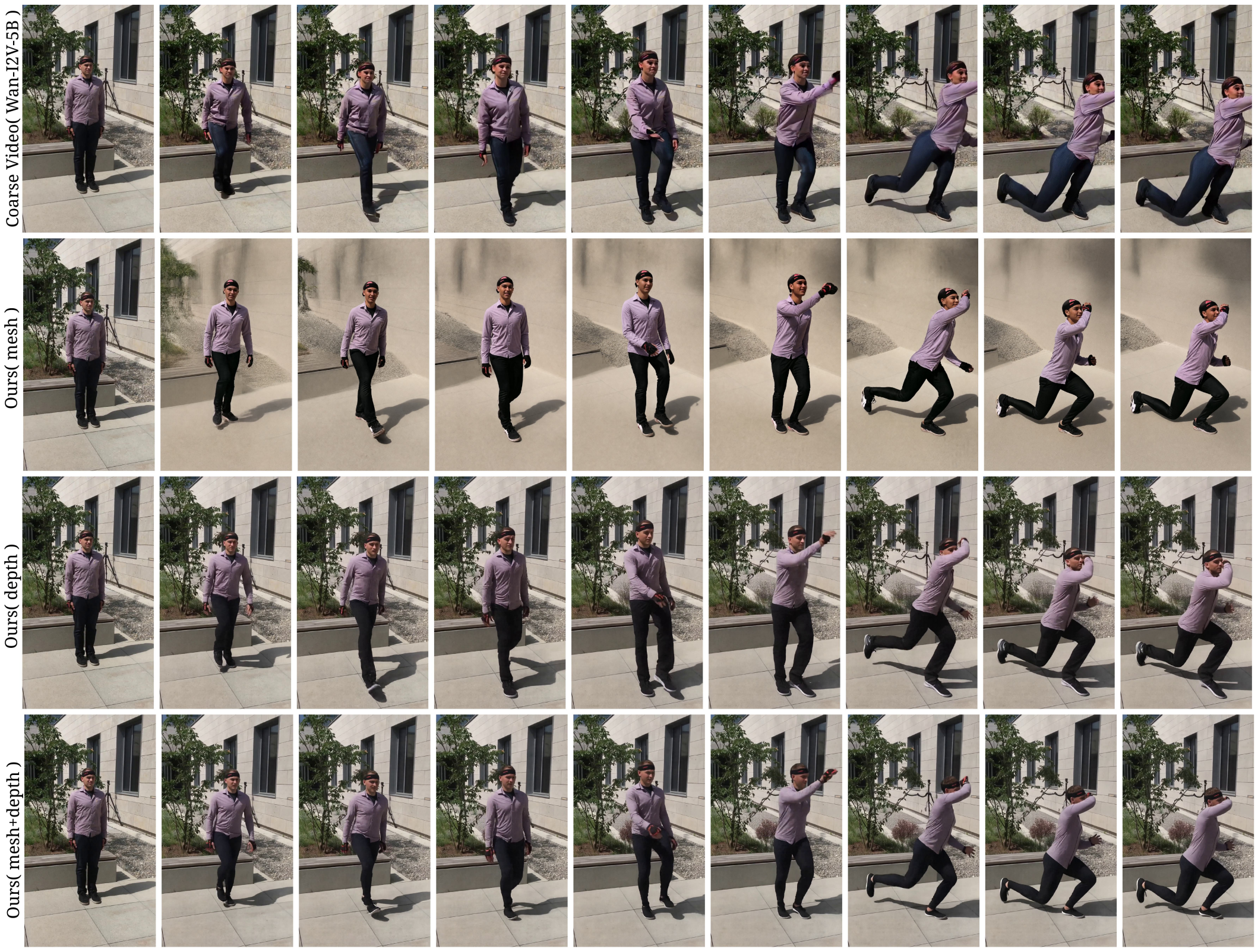}
\caption{
Qualitative ablation of mesh and depth conditioning. Mesh and depth provide complementary constraints on human structure and scene geometry, respectively, while their combination better preserves the motion prior of the coarse video and improves overall generation consistency.
}
\label{fig:comparison}
\end{figure}

\subsection{Additional Experimental Details}
\label{sec:add_experiment}

\paragraph{Training Data.}
We construct our training set based on Motion-X++, resulting in 24k training samples. 
Each sample consists of a target RGB video, a rendered mesh conditioning video, a depth conditioning video, and a motion caption. 
Leveraging the human motion and camera annotations provided by Motion-X++~\citep{zhang2025motion}, we construct the mesh and depth conditions following the procedure described in Sec.~\ref{sec:training_data}. 
All conditioning videos are frame-wise aligned with the target video.

\paragraph{Implementation Details.}
We build ReMoVe upon Wan2.2-TI2V-5B and fine-tune it for conditional video generation. 
During training, the target RGB video, rendered mesh conditioning video, and depth conditioning video are uniformly sampled to 81 frames and resized to $480\times832$. 
The first frame of the target video is additionally provided through the native image-conditioning pathway of the base model. 
The three videos are independently encoded by the pretrained VAE. 
The noisy target-video latent is concatenated with the mesh and depth latents along the channel dimension, resulting in a 144-channel input (expanded from the original 48-channel patch embedding).

We use a LoRA rank of 32 and a learning rate of $1\times10^{-4}$, training for 10 epochs with a gradient accumulation step of 4. 
All experiments are conducted on 8 MetaX C500 GPUs using Accelerate, with gradient checkpointing enabled to reduce memory consumption.

\subsection{Additional Evaluation Details}
\label{sec:add_experiment2}

\paragraph{Evaluation Protocol.}
All methods are evaluated on the same held-out set of 152 challenging human-motion videos from Motion-X++. 
Unless otherwise specified, videos are standardized to 81 frames. 
In the standard generation setting, no ground-truth motion sequence is provided. 
In the controlled ground-truth-motion setting, EchoMotion~\citep{yang2025echomotion}, MTVCrafter~\citep{ding2025mtvcrafter}, and ReMoVe are conditioned on the ground-truth motion annotations from Motion-X++. 
Since MTVCrafter supports generation of only 49 frames, it is evaluated using 49-frame sequences, while the other methods follow the default setting. 
Unlike MTVCrafter and ReMoVe, EchoMotion does not support explicit first-frame or reference-image conditioning in its motion-to-video setting. We therefore follow its native generation protocol without providing an appearance reference.

\paragraph{Metric Definitions.}
VBench~\citep{huang2023vbench} and VBench-2.0~\citep{zheng2025vbench2} metrics include Subject Consistency (SC), Motion Smoothness (MS), Dynamic Degree (DD), Image Quality (IQ), Background Consistency (BC), Aesthetic Quality (AQ), Temporal Flickering (TF), and Human Anatomy (HA). 
These are reported as percentages, where higher is better. 
FVD-16 and FID-VID measure distributional discrepancy between generated and reference videos, where lower is better. 
SSIM, LPIPS, and PSNR measure frame-level reconstruction fidelity to the reference video under the ground-truth-motion setting. Higher values indicate better performance for SSIM and PSNR, whereas lower values are preferred for LPIPS.
Since these three metrics require the generated appearance to match a specific reference, we omit them for methods without appearance-reference conditioning, such as EchoMotion.

\subsection{Additional Baseline Analysis}
\label{sec:add_experiment3}

\paragraph{Human Motion Video Generation.}
The Dynamic Degree results reveal substantially different motion characteristics across the compared methods. CogVideoX-1.5-I2V-5B exhibits very limited motion dynamics: its generated subjects often move slowly with little pose variation, while still suffering from noticeable appearance drift. Such limited temporal variation should also be considered when interpreting its relatively favorable distribution-level scores, since videos with fewer changes across frames are inherently easier to keep visually consistent, without necessarily demonstrating strong capability in complex human motion generation. Wan-TI2V-5B, in contrast, produces substantially stronger dynamics, but challenging actions are often accompanied by excessively rapid or abrupt pose changes that lead to physically unnatural intermediate poses, reflected in its lower Motion Smoothness and Human Anatomy scores relative to ReMoVe. ReMoVe exhibits a more moderate degree of dynamics, but achieves substantially higher Motion Smoothness and Human Anatomy. This behavior is consistent with our design goal: rather than maximizing the amount of motion, the motion refinement process preserves the overall motion tendency of the base model while suppressing abrupt and physically abrupt or incoherent transitions, allowing ReMoVe to maintain meaningful and coherent human motion without relying on unrealistically large frame-to-frame changes.

CoMoVi represents another interesting case. Benefiting from its tightly coupled motion--video co-generation design, it achieves the highest Dynamic Degree among all compared methods and qualitatively generates highly dynamic motions with considerably smoother temporal evolution than the base generators. However, strong motion continuity alone does not guarantee appearance or structural stability: CoMoVi obtains clearly lower Subject Consistency and Human Anatomy, and we observe noticeable identity and appearance drift during large motions. In contrast, ReMoVe complements motion refinement with an explicit rendered mesh condition that anchors human geometry and viewpoint-dependent facial appearance, together with a depth condition that stabilizes scene geometry and background evolution, enabling a better balance among motion coherence, anatomical plausibility, appearance consistency, and overall video quality.

\subsection{Additional Motion-Conditioned Analysis}
\label{sec:add_experiment4}

\paragraph{Motion-Conditioned Video Generation.}
EchoMotion achieves the highest Dynamic Degree (96.05), but performs substantially worse on appearance quality, temporal consistency, distributional fidelity, and reconstruction fidelity. 
As further illustrated in Fig.~\ref{fig:case}, EchoMotion frequently suffers from severe appearance degradation and abrupt visual changes. 
Moreover, since it does not support explicit first-frame or reference-image conditioning, the appearance of the generated subject cannot be directly specified, substantially limiting its applicability in identity-preserving motion-conditioned generation.

MTVCrafter achieves very high consistency-related scores, including SC, MS, BC, TF, and HA, reaching 96.81, 99.71, 97.39, 99.80, and 94.85, respectively. 
However, these scores should be interpreted together with its Dynamic Degree, which is exactly 0.0: not a single one of the 152 generated videos was judged dynamic by VBench's motion threshold. 
As shown qualitatively in Fig.~\ref{fig:case}, its generated videos exhibit only limited motion changes despite being conditioned on large-amplitude ground-truth motions, and noticeable appearance drift occurs when motion does emerge. 
This behavior is also reflected in its substantially worse distributional and reconstruction metrics compared with ReMoVe, including FVD-16 / FID-VID of 83.27 / 47.28 versus 27.71 / 16.27, SSIM of 0.7638 versus 0.7984, LPIPS of 0.1697 versus 0.1214, and PSNR of 19.483 dB versus 22.447 dB.

In contrast, ReMoVe achieves the best IQ, AQ, FVD-16, FID-VID, SSIM, LPIPS, and PSNR while maintaining stable and non-trivial motion generation. 
Qualitatively, Fig.~\ref{fig:case} further shows that ReMoVe better preserves the specified appearance while following challenging human motions with substantially improved structural stability and temporal coherence. 
We attribute this advantage to our explicit mesh-and-depth conditioning design. 
Unlike token-based motion injection, our rendered mesh and depth videos provide spatially aligned structural and scene-level guidance, offering a more direct interface between structured human motion and coherent, appearance-consistent video generation.

\end{document}